\documentclass[acmsmall]{acmart}
\AtBeginDocument{%
  \providecommand\BibTeX{{%
    \normalfont B\kern-0.5em{\scshape i\kern-0.25em b}\kern-0.8em\TeX}}}

\setcopyright{acmcopyright}
\copyrightyear{2018}
\acmYear{2018}
\acmDOI{XXXXXXX.XXXXXXX}

\acmJournal{JACM}
\acmVolume{37}
\acmNumber{4}
\acmArticle{111}
\acmMonth{8}

\usepackage{natbib}
\usepackage[utf8]{inputenc} 
\usepackage[T1]{fontenc}
\usepackage{amsthm}
\usepackage{tikz}
\usepackage[edges]{forest}
\usepackage{hyperref}
\usepackage{graphicx}
\definecolor{hiddendraw}{RGB}{205, 44, 36}
\definecolor{hidden-blue}{RGB}{194,232,247}
\definecolor{hidden-orange}{RGB}{243,202,120}
\definecolor{hidden-yellow}{RGB}{242,244,193}
\theoremstyle{definition}

\newcommand{\header}[1]{\paragraph{{#1}}}

\def\etal{{\em et al.\/}\, }

\begin{document}

\title{A Comprehensive Survey of AI-Generated Content (AIGC): \\A History of Generative AI from GAN to ChatGPT}

\author{Yihan Cao}
\email{yihanc@andrew.cmu.edu}
\authornote{Incoming Ph.D. student at Lehigh University.}
\affiliation{%
  \institution{Lehigh University \& Carnegie Mellon University}
  \city{Pittsburgh}
  \state{PA}
  \country{USA}}

\author{Siyu Li}
\email{applicantlisiyu@hotmail.com}
\affiliation{
    \institution{Lehigh University}
    \city{Bethlehem}
    \state{PA}
    \country{USA}
}
\author{Yixin Liu}
\email{lis221@lehigh.edu}
\affiliation{
    \institution{Lehigh University}
    \city{Bethlehem}
    \state{PA}
    \country{USA}
}
\author{Zhiling Yan}
\email{zhilingyan724@outlook.com}
\affiliation{
    \institution{Lehigh University}
    \city{Bethlehem}
    \state{PA}
    \country{USA}
}
\author{Yutong Dai}
\email{lis221@lehigh.edu}
\affiliation{
    \institution{Lehigh University}
    \city{Bethlehem}
    \state{PA}
    \country{USA}
}

\author{Philip S. Yu}
\affiliation{
  \institution{University of Illinois at Chicago}
  \city{Chicago}
  \state{Illinois}
  \country{USA}
  }
\email{psyu@uic.edu}

\author{Lichao Sun}
\email{lis221@lehigh.edu}
\affiliation{
    \institution{Lehigh University}
    \city{Bethlehem}
    \state{PA}
    \country{USA}
}


\begin{abstract}
Recently, ChatGPT, along with DALL-E-2~\cite{ramesh2021zero} and Codex~\cite{chen2021evaluating},has been gaining significant attention from society. As a result, many individuals have become interested in related resources and are seeking to uncover the background and secrets behind its impressive performance.
In fact, ChatGPT and other Generative AI (GAI) techniques belong to the category of Artificial Intelligence Generated Content (AIGC), which involves the creation of digital content, such as images, music, and natural language, through AI models. The goal of AIGC is to make the content creation process more efficient and accessible, allowing for the production of high-quality content at a faster pace. AIGC is achieved by extracting and understanding intent information from instructions provided by human, and generating the content according to its knowledge and the intent information. In recent years, large-scale models have become increasingly important in AIGC as they provide better intent extraction and thus, improved generation results. With the growth of data and the size of the models, the distribution that the model can learn becomes more comprehensive and closer to reality, leading to more realistic and high-quality content generation. This survey provides a comprehensive review on the history of generative models, and basic components, recent advances in AIGC from unimodal interaction and multimodal interaction. From the perspective of unimodality, we introduce the generation tasks and relative models of text and image. From the perspective of multimodality, we introduce the cross-application between the modalities mentioned above. Finally, we discuss the existing open problems and future challenges in AIGC.
\end{abstract}



\received{20 February 2007}
\received[revised]{12 March 2009}
\received[accepted]{5 June 2009}

\maketitle

\section{Introduction}
\begin{figure*}[h]
    \centering
    \includegraphics[width=0.9\linewidth]{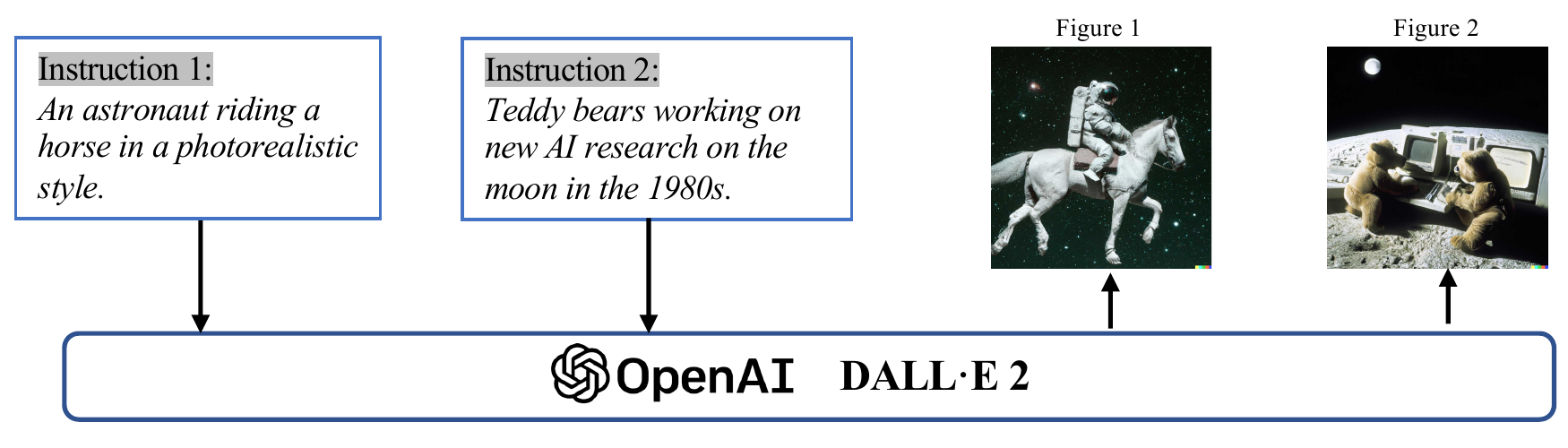}
    \caption{Examples of AIGC in image generation. Text instructions are given to OpenAI DALL-E-2 model, and it generates two images according to the instructions.}
    \label{fig:example}
\end{figure*}

In recent years, Artificial Intelligence Generated Content (AIGC) has gained much attention beyond the computer science community, where the whole society begins to be interested in the various content generation products built by large tech companies~\cite{aigc_importance}, such as ChatGPT~\cite{ChatGPT_2022} and DALL-E-2~\cite{ramesh_hierarchical_2022}.
AIGC refers to content that is generated using advanced Generative AI (GAI) techniques, as opposed to being created by human authors, which can automate the creation of large amounts of content in a short amount of time. For example, ChatGPT is a language model developed by OpenAI for building conversational AI systems, which can efficiently understand and respond to human language inputs in a meaningful way.
In addition, DALL-E-2 is another state-of-the-art GAI model also developed by OpenAI, which is capable of creating unique and high-quality images from textual descriptions in a few minutes, such as "an astronaut riding a horse ina photorealistic style" as shown in Figure~\ref{fig:example}.
As the remarkable achievements in AIGC, many people believe it will be the new era of AI and make significant impacts on the whole world.

Technically, AIGC refers to, given human instructions which could help teach and guide the model to complete the task, 
utilizing GAI algorithms to generate content that satisfies the instruction. 
This generation process usually consists of two steps: extracting intent information from human instructions and generating content according to the extracted intentions. 
However, the paradigm of GAI models containing the above two steps is not entirely novel, as demonstrated by previous studies~\cite{stefanini_show_2021, liang_foundations_2022}. 
The core advancements in recent AIGC compared to prior works are the result of training more sophisticated generative models on larger datasets, using larger foundation model architectures, and having access to extensive computational resources.
For example, the main framework of GPT-3 maintains the same as GPT-2, but the pre-training data size grows from WebText~\cite{Gokaslan2019OpenWeb}(38GB) to CommonCrawl~\cite{brown_language_2020}(570GB after filtering), and the foundation model size grows from 1.5B to 175B.
Therefore, GPT-3 has better generalization ability than GPT-2 on various tasks, such as human intent extraction. 

In addition to the benefits brought by the increase in data volume and computational power, researchers are also exploring ways to integrate new technologies with GAI algorithms. For example, ChatGPT utilizes reinforcement learning from human feedback (RLHF)~\cite{ouyang_training_2022, christiano_deep_2017, stiennon_learning_2020} to determine the most appropriate response for a given instruction, thus improving model's reliability and accuracy over time. This approach allows ChatGPT to better understand human preferences in long dialogues. Meanwhile, in computer vision, stable diffusion~\cite{rombach_high-resolution_2022}, proposed by Stability.AI in 2022, has also shown great success in image generation. Unlike prior methods, generative diffusion models can help generate high-resolution images by controlling the trade-off between exploration and exploitation, resulting in a harmonious combination of diversity in the generated images and similarity to the training data.

By combining these advancements, models have made significant progress in AIGC tasks and have been adopted in various industries, including art~\cite{Anantrasirichai_2021}, advertising~\cite{kietzmann_artificial_2018}, and education~\cite{kandlhofer_2016}.
In the near future, AIGC will continue to be a significant area of research in machine learning.
It is therefore crucial to conduct an extensive review of past research and identify the open problems in this field. 
This survey is the first one that focuses on the core technologies and applications in the field of AIGC. 

\begin{figure*}[t]
    \centering
    \includegraphics[width=1\linewidth]{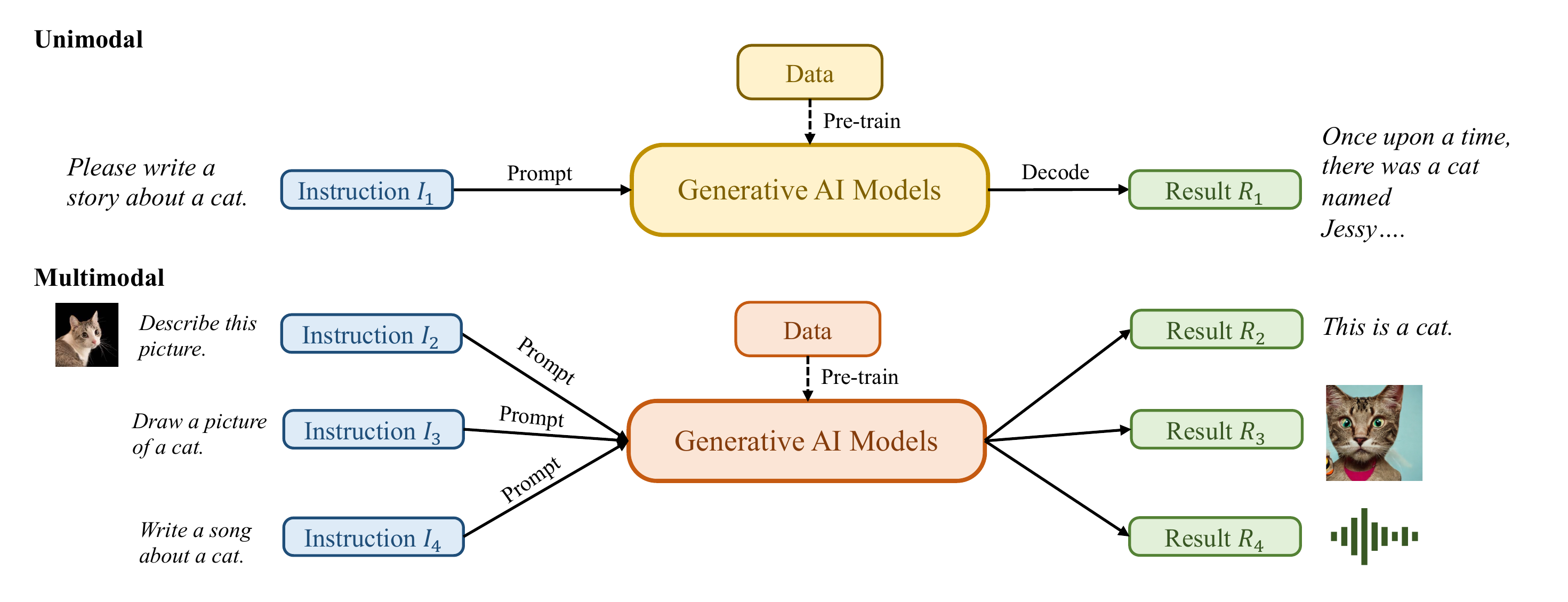}
    \caption{Overview of AIGC. Generally, GAI models can be categorized into two types: unimodal models and multimodal models. Unimodal models receive instructions from the same modality as the generated content modality, whereas multimodal models accept cross-modal instructions and produce results of different modalities.}
    \label{overview}
\end{figure*}

\subsection{Major Contributions}
This is the first comprehensive survey of AIGC that summarizes GAI in the aspects of techniques and applications. Previous surveys have focused on GAI from various angles, including natural language generation~\cite{nlp_generation_survey}, image generation\cite{image_generation_survey}, generation in multimodal machine learning~\cite{liang_foundations_2022, Suzuki_2022}.
However, these prior works only focus on a specific part of AIGC. In this survey, we first provide a review of foundation techniques commonly used in AIGC. Then, we further offer a thorough summary of advanced GAI algorithms, both in terms of unimodal generation and multimodal generation, as shown in Figure~\ref{overview}.
In addition, we examine the applications and potential challenges of AIGC.
Finally, we highlight the open problems and future directions in this field.
In summary, the main contributions of this paper are as follows:

\begin{itemize}
    \item To our best knowledge, we are the first to provide a formal definition and a thorough survey for AIGC and AI-enhanced generation process.
    \item We review the history, foundation techniques of AIGC and conduct a comprehensive analysis of recent advances in GAI tasks and models from the perspective of unimodal generation and multimodal generation.
    \item We discuss the main challenges facing AIGC and future research trends confronting AIGC.
\end{itemize}

\subsection{Organization}
The rest of the survey is organized as follows. 
Section~\ref{sec:2} reviews the history of AIGC mainly from the view of vision and language modalities. 
Section~\ref{sec:3}
introduces the basic components that are widely used in nowadays GAI model training.
Section~\ref{sec:4} summarizes recent advances of GAI models, among which, Section~\ref{sec:4.1} reviews the advances from unimodal perspective and Section~\ref{sec:4.2} reviews the advances from the perspective of multimodal generation. Among multimodal generation, we introduce vision language models, text audio models, text graph models and text code models.
Section~\ref{sec:5} and Section~\ref{sec:6} introduce the applications of GAI models in AIGC and some other important research that are related to this area.
Furthermore, Sections~\ref{sec:7},~\ref{sec:8} reveal the risk, open problems and future directions of AIGC technologies.
Finally, we conclude our research in~\ref{sec:9}.

\section{History of Generative AI}\label{sec:2}
Generative models have a long history in artificial intelligence, dating back to the 1950s with the development of Hidden Markov Models (HMMs)~\cite{knill_hidden_1997} and Gaussian Mixture Models (GMMs)~\cite{reynolds2009gaussian}. These models generated sequential data such as speech and time series. However, it wasn't until the advent of deep learning that generative models saw significant improvements in performance.

In early years of deep generative models, different areas do not have much overlap in general.
In natural language processing (NLP), 
a traditional method to generate sentences is to learn word distribution using N-gram language modeling~\cite{ngram} and then search for the best sequence. However, this method cannot effectively adapt to long sentences. To solve this problem, 
recurrent neural networks (RNNs)~\cite{mikolov2010recurrent} were later introduced for language modeling tasks , allowing for modeling relatively long dependency. 
This was followed by the development of Long Short-Term Memory (LSTM)~\cite{graves2012long} and Gated Recurrent Unit (GRU)~\cite{dey2017gate}, which leveraged gating mechanism to control memory during training. These methods are capable of attending to around 200 tokens in a sample~\cite{kwal2018analysis}, which marks a significant improvement compared to N-gram language models. 

Meanwhile, 
in computer vision (CV), 
before the advent of deep learning-based methods, traditional image generation algorithms used techniques such as texture synthesis~\cite{efros1999texture} and texture mapping~\cite{heckbert1986survey}. These algorithms were based on hand-designed features, and were limited in their ability to generate complex and diverse images.
In 2014, Generative Adversarial Networks (GANs)~\cite{goodfellow2014gan} was first proposed, which was a significant milestone in this area, due to its impressive results in various applications. 
Variational Autoencoders (VAEs)~\cite{kingma2013auto} and other methods like diffusion generative models~\cite{song2019DSM} 
have also been developed for more fine-grained control over the image generation process and the ability to generate high-quality images.

The advancement of generative models in various domains has followed different paths, but eventually, the intersection emerged: the transformer architecture~\cite{NIPS2017_transformer}.
Introduced by Vaswani et al. for NLP tasks in 2017, Transformer has later been applied in CV and then become the dominant backbone for many generative models in various domains~\cite{brown_language_2020, dalle, lewis2019bart}.
In the field of NLP, many prominent large language models, e.g., BERT and GPT, adopt the transformer architecture as their primary building block, offering advantages over previous building blocks, i.e., LSTM and GRU.
In CV, Vision Transformer (ViT)~\cite{dosovitskiy2020image} and Swin Transformer~\cite{liu2021swin} later takes this concept even further by combining the transformer architecture with visual components, allowing it to be applied to image based downstreams. 
Except for the improvement that transformer brought to individual modalities, this intersection also enabled models from different domains to be fused together for multimodal tasks.
One such example of multimodal models is CLIP~\cite{clip2021}. CLIP is a joint vision-language model that combines the transformer architecture with visual components, allowing it to be trained on a massive amount of text and image data. Since it combines visual and language knowledge during pre-training, it can also be used as image encoders in multimodal prompting for generation. 
In all, the emergence of transformer based models revolutionized AI generation and led to the possibility of large-scale training.

In recent years, researchers have also begun to introduce new techniques based on these models. 
For instance, in NLP, instead of fine-tuning, people sometimes prefer few-shot prompting~\cite{in_context}, which refers to including a few examples selected from the dataset in the prompt, to help the model better understand task requirements.
And in visual language, researchers often combine modality-specific models with self-supervised contrastive learning objectives to provide more robust representations.

In the future, as AIGC becomes increasingly important, more and more technologies shall be introduced, empowering this area with vitality.
\begin{figure}[t]
    \centering
    \includegraphics[width=0.9\linewidth]{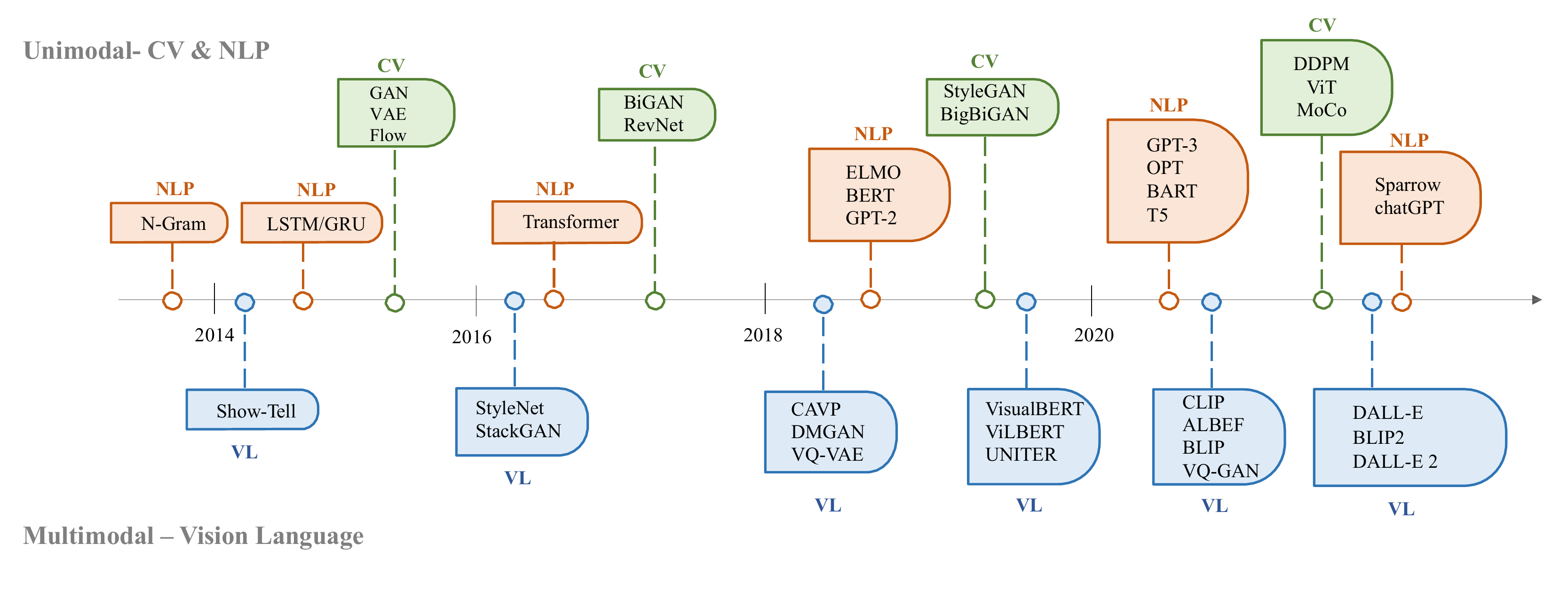}
    \caption{The history of Generative AI in CV, NLP and VL.}
    \label{history}
\end{figure}

\section{Foundations for AIGC}\label{sec:3}
In this section, we introduce foundation models that are commonly used in AIGC. 
\subsection{Foundation Model}


\subsubsection{Transformer}
Transformer is the backbone architecture for many state-of-the-art models, such as GPT-3~\cite{brown_language_2020}, DALL-E-2~\cite{ramesh_hierarchical_2022}, Codex~\cite{chen2021evaluating}, and Gopher~\cite{rae2021scaling}. It was first proposed to solve the limitations of traditional models such as RNNs in handling variable-length sequences and context-awareness. 
Transformer architecture is mainly based on a self-attention mechanism that allows the model to attend to different parts in a input sequence.
Transformer consists of an encoder and a decoder. The encoder takes in the input sequence and generates hidden representations, while the decoder takes in the hidden representation and generates output sequence. Each layer of the encoder and decoder consists of a multi-head attention and a feed-forward neural network. The multi-head attention is the core component of Transformer, which learns to assign different weights to tokens according their relevance.
This information routing method allows the model to be better at handling long term dependency, hence, improving the performance in a wide range of NLP tasks.
Another advantage of transformer is that its architecture makes it highly parallelizable, and allows data to trump inductive biases~\cite{elhage2021mathematical}. This property makes transformer well-suited for large-scale pre-training, enabling transformer based models to become adaptable to different downstream tasks. 
\subsubsection{Pre-trained Language Models}
\begin{figure*}[t]
    \centering
    \includegraphics[width=0.9\linewidth]{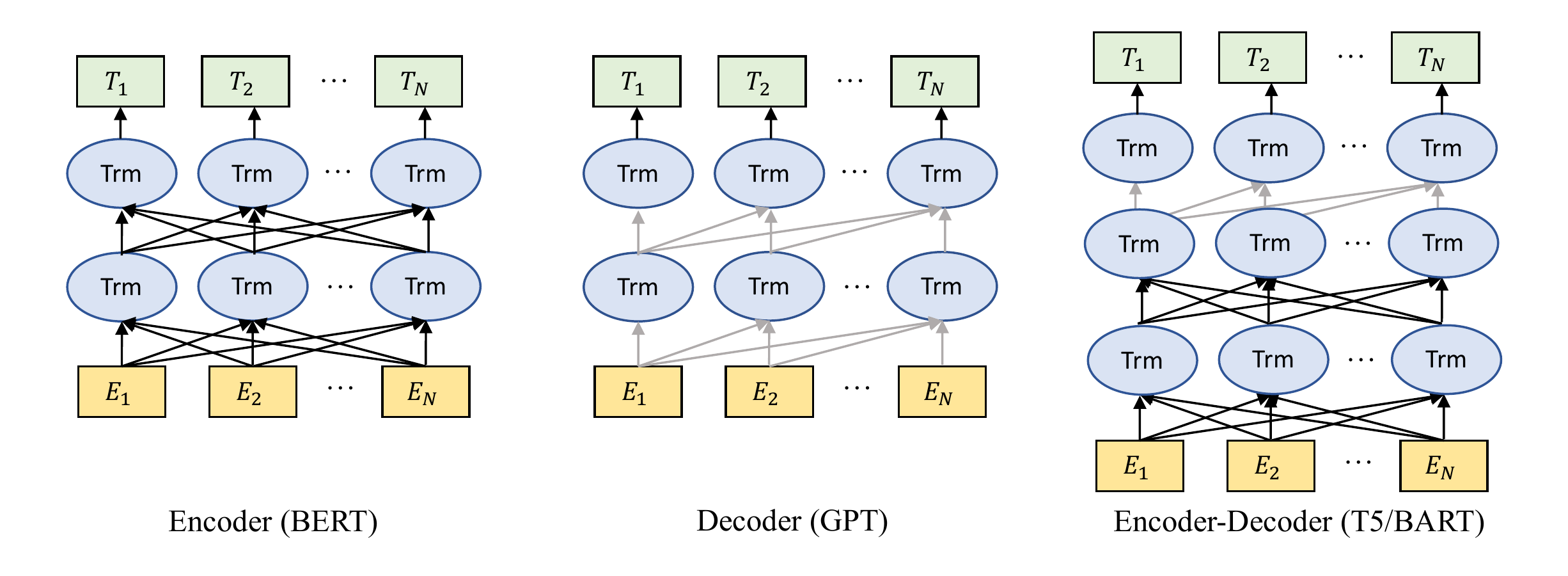}
    \caption{Categories of pre-trained LLMs. Black line represents information flow in bidirectional models, while gray line representas left-to-right information flow. Encoder models, e.g. BERT, are trained with context-aware objectives. Decoder models, e.g. GPT, are trained with autoregressive objectives. Encoder-decoder models, e.g. T5 and BART, combines the two, which use context-aware structures as encoders and left-to-right structures as decoders.}
    \label{Fig:pretrain}
\end{figure*}
Since the introduction of the Transformer architecture, it has become the dominant choice in natural language processing due to its parallelism and learning capabilities. 
Generally, these transformer based pre-trained language models can be commonly classified into two types based on their training tasks: autoregressive language modeling and masked language modeling~\cite{qiu2020pre}. 
Given a sentence, which is composed of several tokens, the objective of masked language modeling, e.g., BERT~\cite{devlin2018bert} and RoBERTa~\cite{liu2019roberta}, refers to predicting the probability of a masked token given context information. 
The most notable example of masked language modeling is BERT~\cite{devlin2018bert}, which includes masked language modeling and next sentence prediction tasks. 
RoBERTa~\cite{liu2019roberta}, which uses the same architecture as BERT, improves its performance by increasing the amount of pre-training data and incorporating more challenging pre-training objectives. XL-Net~\cite{yang2019xlnet}, which is also based on BERT, incorporates permutation operations to change the prediction order for each training iteration, allowing the model to learn more information across tokens. While autoregressive language modeling, e.g., GPT-3~\cite{brown_language_2020} and OPT~\cite{opt}, is to model the probability of the next token given previous tokens, hence, left-to-right language modeling. 
Different from masked language models, autoregressive models are more suitable for generative tasks.
We will introduce more about autoregressive models in Section \ref{sec:glm}.


\subsection{Reinforcement Learning from Human Feedback} 
Despite being trained on large-scale data, the AIGC may not always produce output that aligns with the user's intent, which includes considerations of usefulness and truthfulness. In order to better align AIGC output with human preferences, reinforcement learning from human feedback (RLHF) has been applied to fine-tune models in various applications such as Sparrow, InstructGPT, and ChatGPT \cite{glaese2022improving, ouyang_training_2022}.

Typically, the whole pipeline of RLHF includes the following three steps: \textit{pre-training, reward learning, and fine-tuning with reinforcement learning}. First, a language model $\theta_0$ is pre-trained on large-scale datasets as an initial language model. Since the {\it (prompt-answer)} pair given by $\theta_0$ might not align with human purposes, in the second step we train a reward model to encode the diversified and complex human preference. Specifically, given the same prompt $x$, different generated answers $\{y_1,y_2, \cdots, y_3\}$ are evaluated by humans in a pairwise manner. The pairwise comparison relationships are later transferred to pointwise reward scalars, $\{r_1,r_2, \cdots, r_3\}$, using an algorithm such as ELO \cite{coulom2008whole}. In the final step, the language model $\theta$ is fine-tuned to maximize the learned reward function using reinforcement learning. To stabilize the RL training, Proximal Policy Optimization (PPO) is often used as the RL algorithm. In each episode of RL training, an empirically-estimated KL penalty term is considered to prevent the model from outputting something peculiar to trick the reward model. Specifically, the total reward $r_{total}$ at each step is given by $r_{total}(x,y)=r_{RM}(x,y)-\lambda_{\mathrm{KL}} D_{\mathrm{KL}}\left(\pi_\theta | {\pi_{\theta_0}}\right)$, where $r_{RM}$ is the learned reward model, $D_{\mathrm{KL}}$ is the KL penalty term, and $\pi_\cdot$ is the trained policy. For more details on RLHF, please refer to \cite{christiano2017deep}.


Although RLHF has shown promising results by incorporating fluency, progress in this field is impeded by a lack of publicly available benchmarks and implementation resources, leading to a perception that RL is a challenging approach for NLP. To address this issue, an open-source library named RL4LMs~\cite{ramamurthy2022reinforcement} has recently been introduced, consisting of building blocks for fine-tuning and evaluating RL algorithms on LM-based generation.

Beyond human feedback, the latest dialogue agent, Claude, favors Constitutional AI \cite{bai2022constitutional}, where the reward model is learned via RL from AI Feedback (RLAIF). Both the critiques and the AI feedback are guided by a small set of principles drawn from a ``constitution'', which is the only thing provided by humans in Claude. The AI feedback focuses on controlling the outputs to be less harmful by explaining its objections to dangerous queries. Moreover, recently a preliminary theoretical analysis of the RLAIF \cite{zhu2023principled} justifies the empirical success of RLHF and provides new insights for specialized RLHF algorithm design for language models.



\subsection{Computing}
\begin{figure*}[t]
    \centering
    \includegraphics[width=\linewidth]{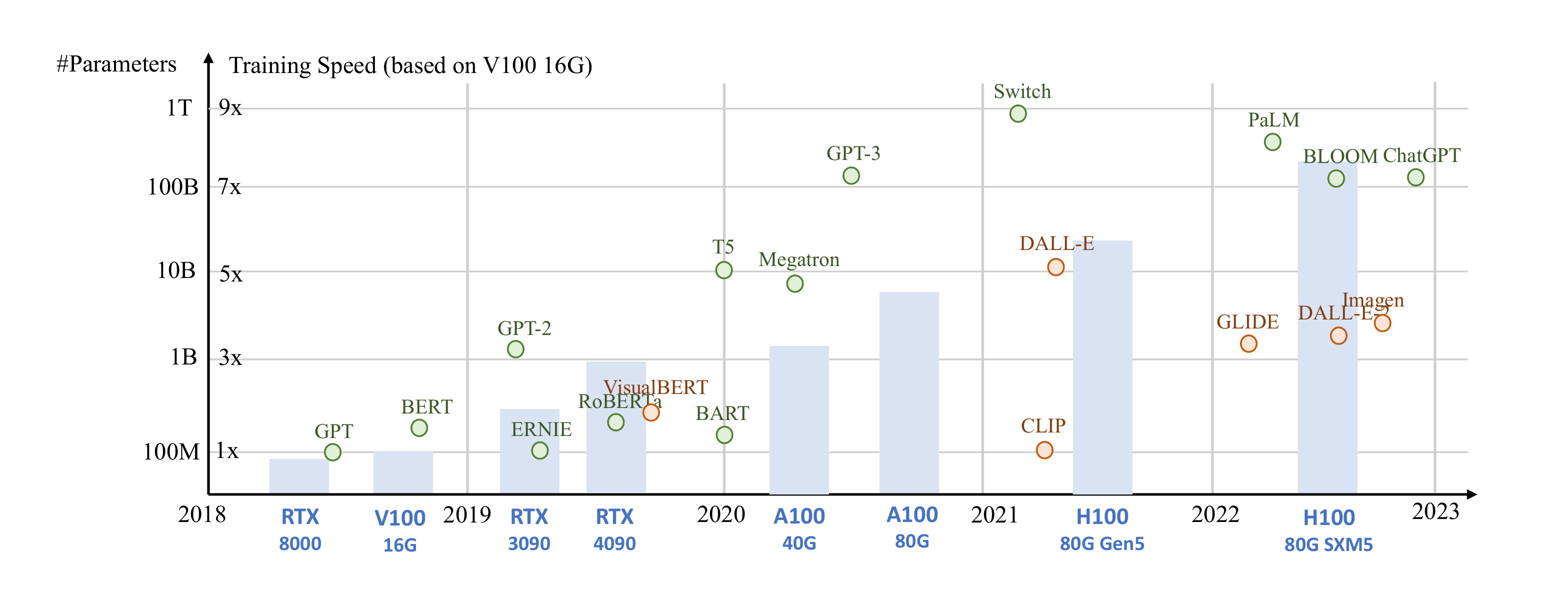}
    \caption{Statistics of model size~\protect\cite{amatriain2023transformer} and training speed~\protect\footnotemark across different models and computing devices.}
    \label{Fig:computing}
\end{figure*}
\footnotetext{\url{https://lambdalabs.com/gpu-benchmarks}}
\subsubsection{Hardware}
In recent years, there have been significant hardware advancements that have facilitated the training of large-scale models. In the past, training a large neural network using CPUs could take several days or even weeks. However, with the emergence of more powerful computing resources, this process has been accelerated by several orders of magnitude. For instance, the NVIDIA A100 GPU achieves seven times faster during BERT-large inference compared to the V100 and 11 times faster than the T4\footnote{\url{https://www.nvidia.com/content/dam/en-zz/Solutions/Data-Center/a100/pdf/nvidia-a100-datasheet-nvidia-us-2188504-web.pdf}}. Additionally, Google's Tensor Processing Units (TPUs), which are designed specifically for deep learning, offer even higher computing performance compared to the current generation of A100 GPUs\footnote{\url{https://cloud.google.com/blog/products/ai-machine-learning/google-wins-mlperf-benchmarks-with-tpu-v4}}. This rapid progress in computing power has significantly increased the efficiency of AI model training and opened up new possibilities for developing large and complex models.

\subsubsection{Distributed training}
Another significant improvement is distributed training. In traditional machine learning, training is typically performed on a single machine using a single processor. This approach can work well for small datasets and models, but it becomes impractical when dealing with large datasets and complex models. In distributed training, the training workload is split among multiple processors or machines, allowing the model to be trained much faster. Some companies have also released frameworks that simplify the process of distributed training on deep learning stacks~\cite{horovod, rasley2020deepspeed, dai2022bigdl}. These frameworks provide tools and APIs that allow developers to easily distribute their training workloads across multiple processors or machines, without having to manage the underlying infrastructure.

\subsubsection{Cloud computing}
Cloud computing has also played a vital role in training large-scale models. Previously, models are often trained locally. Now with the cloud computing services like AWS and Azure providing access to powerful computing resources, deep learning researchers and practitioners could spin up large clusters of GPUs or TPUs as needed for training large-scale models. Overall, these advancements have enabled the development of more complex and accurate models, unlocking new possibilities in various areas of AI research and applications.

\section{Generative AI}\label{sec:4}
\subsection{Unimodal Models} \label{sec:4.1} 
In this section, we will introduce state-of-the-art unimodal generative models. These models are designed to accept a specific raw data modality as input, such as text or images, and then generate predictions in the same modality as the input.
We will discuss some of the most promising approaches and techniques used in these models, including 
generative language models, e.g., GPT-3~\cite{brown_language_2020}, BART~\cite{lewis2019bart}, T5~\cite{raffel2020exploring}, and generative vision models, e.g., GAN~\cite{goodfellow2014gan}, VAE~\cite{kingma2013auto}, and normalizng flow~\cite{dinh2014nice}.

\subsubsection{Generative Language Models}\label{sec:glm}
Generative language models (GLMs) are a type of NLP models that are trained to generate readable human language based on patterns and structures in input data that they have been exposed to. These models can be used for a wide range of NLP tasks such as 
dialogue systems~\cite{ni2022recent}, 
, translation~\cite{yang2020survey} and question answering~\cite{zhu2021retrieving}.

\begin{figure*}[t]
    \centering
    \includegraphics[width=0.9\linewidth]{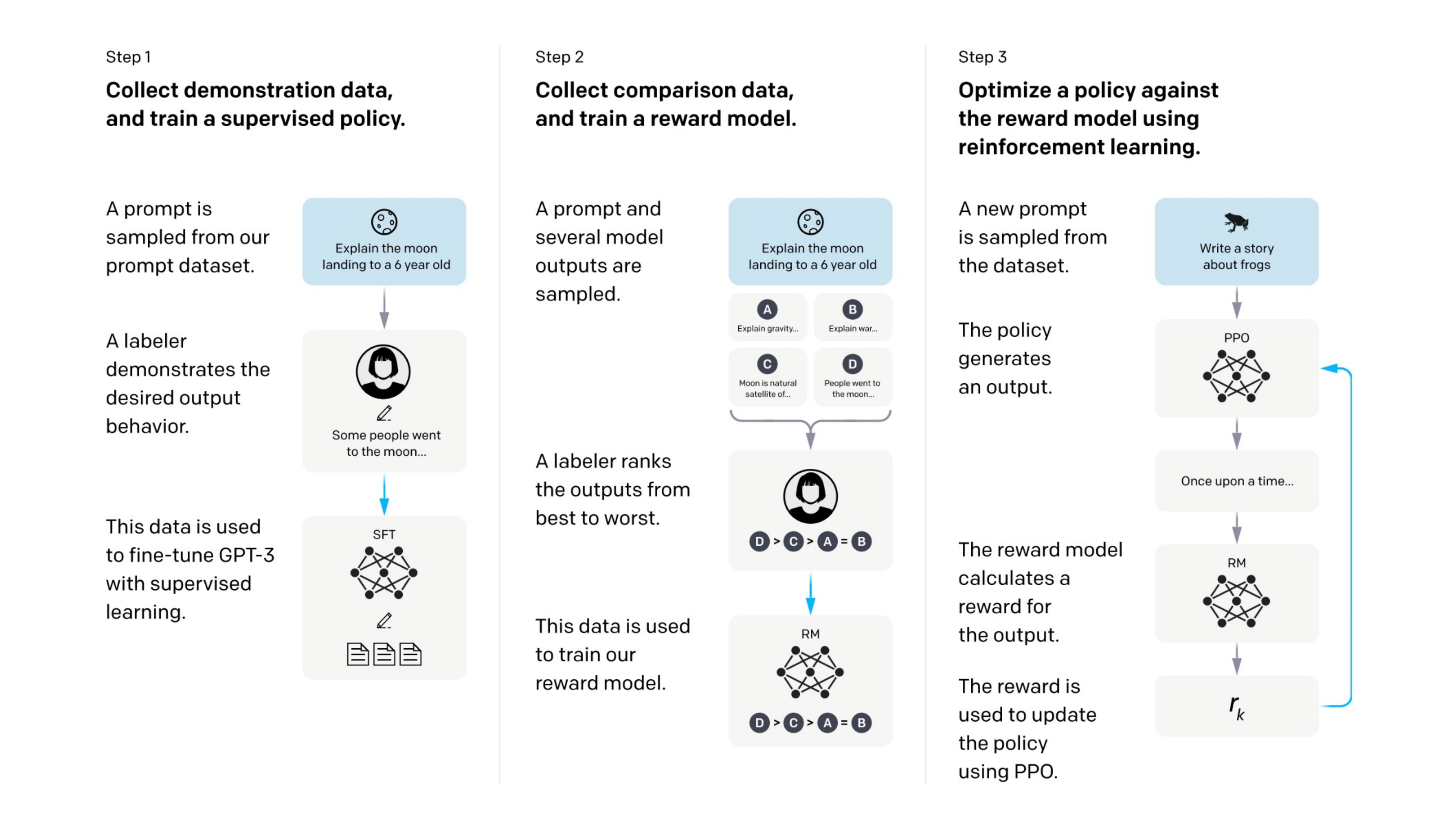}
    \caption{The architecture of InstructGPT~\cite{ouyang_training_2022}. First, demonstration data are collected with human labelers and is used to fine-tune GPT-3. Then prompts and corresponding answers are sampled from the language model and human labelers will rank the answers from best to worst. This data is used to train a reward model. Finally, with the trained reward model, the language model could be optimized according to the preference of human labelers.}
    \label{Fig:InstructGPT}
\end{figure*}

Recently, The use of pre-trained language models has emerged as the prevailing technique in the domain of NLP.
Generally, current state-of-the-art pre-trained language models could be categorized as masked language models (encoders), autoregressive language models (decoders) and encoder-decoder language models, as shown in Figure~\ref{Fig:pretrain}.
Decoder models are widely used for text generation, while encoder models are mainly applied to classification tasks. By combining the strengths of both structures, encoder-decoder models can leverage both context information and autoregressive properties to improve performance across a variety of tasks. The primary focus of this survey is on generative models. In the following sections, we will delve into recent advancements in decoder and encoder-decoder architectures.

\header{\textbf{Decoder Models.}}
One of the most prominent examples of autoregressive decoder-based language models is GPT~\cite{Radford2018ImprovingLU}, which is a transformer-based model that utilizes self-attention mechanisms to process all words in a sequence simultaneously. 
GPT is trained on next word prediction task based on previous words, allowing it to generate coherent text.
Subsequently, GPT-2~\cite{Radford2019LanguageMA} and GPT-3~\cite{brown_language_2020} maintains the autoregressive left-to-right training method, while scaling up model parameters and leveraging diverse datasets beyond basic web text, achieving state-of-the-art results on numerous datasets. 
Gopher~\cite{rae2021scaling} uses a GPT-like structure but replace LayerNorm~\cite{ba2016layer} with RSNorm, where a residual connection is added to the original layernorm structure to maintain the information.
In addition to enhancing the normalization function, several other studies have concentrated on optimizing the attention mechanism. BLOOM~\cite{scao2022bloom} shares the same structure as GPT-3 but instead of using sparse attention, BLOOM uses a full attention network, which is better suited for modeling long dependencies.
~\cite{megatron} proposes Megatron, which extends commonly used architectures like GPT-3, BERT and T5 with distributed training objectives to process large amount of data.
This method is also later adopted by MT-NLG~\cite{mt-nlg} and OPT~\cite{opt}.
Except for the advancements in model architecture and pre-training tasks, there has also been significant efforts put into improving the fine-tuning process for language models.
For example, InstructGPT~\cite{ouyang_training_2022} takes advantage of pre-trained GPT-3 and uses RLHF for fine-tuning, allowing the model to learn preference according to ranking feedback labeled by human.

\header{\textbf{Encoder-Decoder Models.}}
One of the main encoder-decoder methods is Text-to-Text Transfer Transformer (T5)~\cite{raffel2020exploring}, which combines transformer-based encoders and decoders together for pre-training. T5 employs a "text-to-text" approach, which means that it transforms both the input and output data into a standardized text format. This allows T5 to be trained on a wide range of NLP tasks, such as machine translation, question-answering, summarization, and more, using the same model architecture.
Switch Transformer~\cite{fedus2021switch}, as stated in its name, utilizes "switching", which refers to a simplified MoE routing algorithm, for parallelized training on T5. This model successfully obtained larger scale and better performance with the same computational resources compared to the base model.
Another widely-used method that improves upon T5 is ExT5~\cite{aribandi2021ext5}, which is proposed by Google in 2021, extending the scale of previous T5 model.
Compared to T5, ExT5 is continue pre-trained on C4 and ExMix, which is a combinition of 107 supervised NLP tasks across diverse domains. 
Another widely used encoder-decoder method is BART~\cite{lewis2019bart}, which blends the bidirectional encoder from BERT and the autoregressive decoder from GPT, allowing it to leverage the bidirectional modeling abilities of the encoder while retaining the autoregressive properties for generation tasks. 
HTLM~\cite{html} leverages BART denoising objectives for modeling hyper-text language, which contains valuable information regarding document-level structure. This model also achieves state-of-the-art performance on zero-shot learning on various generation tasks.
While DQ-BART~\cite{li2022dq}, instead, aims at compressing BART into a smaller model using distillation and quantization, which achieves the BART original performance on various downstream tasks.

\subsubsection{Vision Generative Models}
\begin{figure*}[t]
    \centering
    \includegraphics[width=0.9\linewidth]{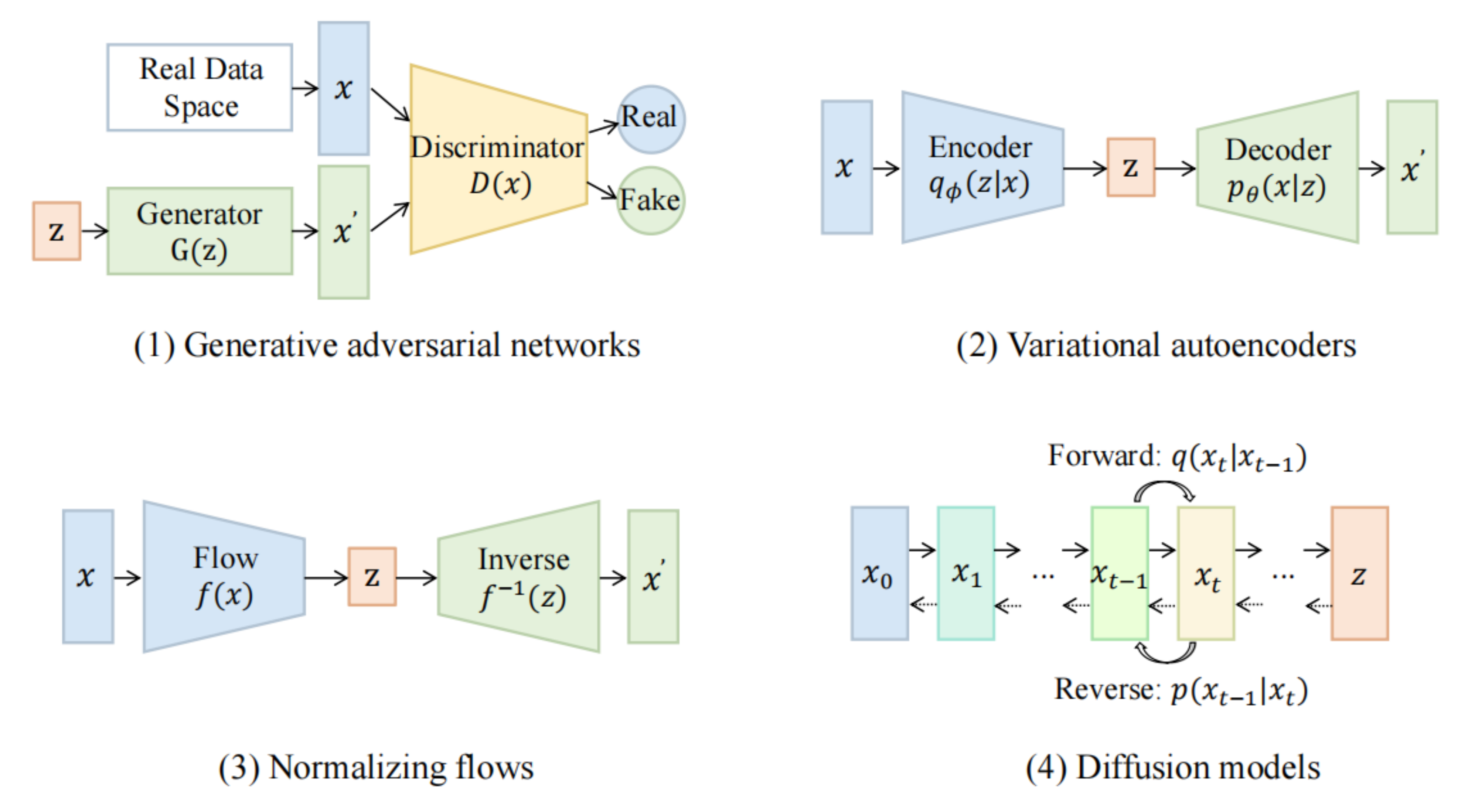}
    \caption{Categories of vision generative models.}
    \label{fig:vision_unimodal}
\end{figure*}

\paragraph{\textbf{GAN}} Generative Adversarial Networks (GANs) have gained popularity in the field of image generation research. GANs consist of two parts, a generator and a discriminator. The generator attempts to learn the distribution of real examples in order to generate new data, while the discriminator determines whether the input is from the real data space or not.

\textit{Structure.} The structure of the generator and the discriminator highly influence GAN's training stability and performance. LAPGAN~\cite{denton2015deep} generates high-quality images in a coarse-to-fine fashion using a cascade of convolutional networks within a Laplacian pyramid framework ~\cite{burt1987laplacian}. A. Radford et al.~\cite{radford2015unsupervised} propose DCGANs structure, a class of CNNs with architectural constraints, as a powerful solution for unsupervised learning. Progressive GAN ~\cite{karras2017progressive} progressively grows the generator and discriminator, starting from low resolution and adding layers to model finer details, resulting in faster and more stable training and producing high-quality images. As traditional convolutional GANs generate high-resolution details based only on spatially local points in lower-resolution feature maps, SAGAN ~\cite{zhang2019self} introduces attention-driven, long-range dependency modeling and spectral normalization for improved training dynamics. 
In addition, generating high-resolution and diverse samples from complex datasets remains a challenge. To address this, BigGAN~\cite{brock2018large} is proposed as a large scale TPU implementation of GANs. StyleGAN~\cite{karras2019style} improves GANs by separating high-level attributes and variations, allowing for intuitive control and better performance in terms of quality metrics, interpolation, and disentanglement.~\cite{donahue2016adversarial, ulyanov2018takes} focus on inverse mapping - projecting data back into the latent space, resulting in a useful feature representation for auxiliary discrimination tasks. 
To address mode collapse and improve the generative model, both the D2GAN~\cite{nguyen2017dual} and GMAN~\cite{durugkar2016generative} methods extend the traditional GANs by combining extra discriminators. MGAN~\cite{hoang2017multi} and MAD-GAN~\cite{ghosh2018multi} address the mode collapse problem by incorporating multiple generators and one discriminator. CoGAN~\cite{liu2016coupled} is composed of a pair of GANs with a weight-sharing constraint, allowing for learning the joint distribution from separate marginal distributions without requiring corresponding images in the training set.
    
\textit{Representative variants.}
As the latent vector $z$ of the generator is highly unstructured, InfoGAN~\cite{chen2016infogan} proposes another latent code $c$ to extract the significant structured features of the actual data space. In CGANs~\cite{mirza2014conditional, lu2018attribute, mao2019mode}, the generator and discriminator are conditioned on additional information, such as class labels or data from other modalities, to generate samples that are conditioned on specific attributes. f-GAN~\cite{nowozin2016f} allows for the use of any f-divergence as the objective function for training the generative model. The choice of f-divergence provides a flexible framework for controlling the trade-off between the quality of the generated samples and the difficulty of training the model. 
    
\textit{Objective function.} The goal of generative models is to match the real data distribution. WGAN~\cite{gulrajani2017improved} and LS-GAN~\cite{qi2020loss, mao2017least} aim to regularize the loss function with a Lipschitz regularity condition on the density of real data in order to better generalize and produce realistic new data. \cite{miyato2018spectral} is a weight normalization technique proposed to stabilize the training of the discriminator in GANs. Che et al.~\cite{che2016mode} regularize the objective, which can stabilize the training of GAN models. UGAN~\cite{metz2016unrolled} stabilizes training of GANs by defining the generator objective with respect to an unrolled optimization of the discriminator. \cite{jolicoeur2018relativistic} makes discriminator relativistic by sampling from real/generated data pairs to improve stability and coverage of the data distribution generated by the generator.

\paragraph{\textbf{VAE}} Following variational bayes inference~\cite{fox2012tutorial}, Variational Autoencoders (VAE) are generative models that attempt to reflect data to a probabilistic distribution and learn reconstruction that is close to its original input. 

\textit{Complex priors.} Rewriting the variational evidence lower bound objective (ELBO) of variational autoencoders contributes to improve the variational bounds~\cite{hoffman2016elbo}. Since the true aggregate posterior is intractable, VampPrior~\cite{tomczak2018vae} introduces a variational mixture of posteriors priors conditioned on learnable pseudo-inputs. \cite{maaloe2019biva, vahdat2020nvae, wu2021greedy} propose skip connections around the stochastic sampling process to capture different aspects of the data distribution. 

\textit{Regularized Autoencoders.} \cite{ghosh2019variational, ramesh2021zero, van2017neural} introduce regularisation to the latent space of the encoder and lead to a smooth and representative latent space without conforming to an arbitrarily chosen prior. \cite{razavi2019generating} propose a multi-scale hierarchical organization to model larger images.

\paragraph{\textbf{Flow}} A Normalizing Flow is a distribution transformation from simple to complex by a sequence of invertible and differentiable mappings. 

\textit{Coupling and autoregressive flows.} A non-linear deterministic transformation of the data is learned through a coupling method in \cite{dinh2014nice} to make the transformed data conform to a factorized distribution. Dinh et al.~\cite{dinh2016density} proposes multi-scale flow to gradually introduce dimensions to the distribution in the generative direction. A more flexible generalisation of coupling layers is the autoregressive flow~\cite{papamakarios2017masked, huang2018neural, de2020block}, which permits parallel density estimation as a universal approximator.

\textit{Convolutional and Residual Flows.} Zheng et al.~\cite{zheng2017convolutional} used 1D convolutions (ConvFlow) and Hoogeboom et al.~\cite{hoogeboom2019emerging} have provided a more general solution for modelling d×d convolutions. They exploited the triangular structure to improve the interaction among inputs and efficiently compute the determinant. RevNets~\cite{gomez2017reversible} and iRevNets~\cite{jacobsen2018revnet} are the first to build a reversible network architecture based on residual connections, which alleviate the vanishing gradients problem. In addition, the residual connections can be viewed as discretizations of a first order ordinary differential equation (ODE)~\cite{haber2018learning} to improve parameter efficiency.

\paragraph{\textbf{Diffusion}} The Generative Diffusion Model (GDM) is a cutting-edge class of generative models based on probability, which demonstrates state-of-the-art results in the field of computer vision. It works by progressively corrupting data with multiple-level noise perturbations and then learning to reverse this process for sample generation.

\textit{Model Formulations.} Diffusion Models are mainly formulated into three categories. DDPM \cite{ho2020denoising} applies two Markov chains respectively to progressively corrupt data with Gaussian noise and reverse the forward diffusion process by learning Markov transition kernels.  Score-based generative models (SGMs) directly work on the gradient of log density of data a.k.a score function. NCSN \cite{song2019DSM} perturbs data with multi-scale intensifying noise and jointly estimates score function of all such noisy data distribution by a neural network conditioned on all noise levels. It enjoys flexible sampling due to the completely decoupled training and inference steps. Score SDE \cite{song2020score} generalizes previous two formulations into continuous settings, where noise perturbations and denoising processes are solutions to stochastic differential equations. It is proved that probability flow ODE could also be used to model the reverse process.  
    
\textit{Training Enhancement.} Training enhancement aims to improve sampling by introducing prior knowledge from another pre-trained model or extra trainable hyper-parameters. Inspired from the idea of knowledge distillation, Salimans et al. \cite{salimans2022progressive} propose to progressively distill knowledge from a pre-trained complicated teacher model to a faster student model, which could cut sampling steps in half. TDPM \cite{zheng2022truncated} and ES-DDPM \cite{lyu2022accelerating} improve sampling speed by truncating the diffusion process with early stop. To generate sample from reverse process initialized by a non-Gaussian distribution, another pre-trained generative model such as VAE or GAN is introduced to approximate such distribution. Franzese et al. \cite{2206.05173} formulate the number of training steps as a variable to realize an optimal trade-off. Improved DDPM \cite{nichol2021improved} first introduces noise scale tuning by adding noise scale term into loss function.Meanwhile, San Romans et al \cite{san2021noise} introduce a noise prediction network to enable noise schedule adjustment step-by-step. Such noise schedule learning improves reconstruction by efficiently guiding the random walk of noise during training and inference.
    
\textit{Efficient Training-free Sampling.} Instead of additional training, training-free sampling directly reduce the number of discretized time steps while minimizing discretization errors. Under same training objective, DDIM \cite{song2020denoising} generalizes DDPM to a class of non-Markovian diffusion process and introduces jump-step acceleration. This could provide shorter generative Markov chains. Analytic-DPM \cite{bao2022analytic} provides more efficient inference by estimating the analytic form of optimal model reverse variance and KL-divergence w.r.t its score function. There are also works \cite{watson2021learning,watson2022learning} which directly figure out optimal sampling trajectories via dynamic programming.
    
\textit{Noise Distribution.} The distribution of noise perturbations is an essential part of diffusion models and most of them are Gaussian. Meanwhile, fitting such distribution with more degrees of freedom could benefit performance. Nachmani et al. \cite{2106.07582} prove that Gamma distribution could improve image and speech generation and a mixture of Gaussian distribution also outperforms a single distribution.Furthermore, cold diffusion \cite{bansal2022cold} proposes a more generalized conclusion that noise can be set to any distribution as the generative behavior of diffusion model is not strongly dependent on the choice of noise distribution. Apart from noise perturbation, CCDF \cite{2112.05146} shows it is unnecessary to initialize from Gaussian distribution and it could reduce sampling steps with a simple forward diffusion but better noise initialization. 
    
\textit{Mixed Modeling.} Mixed-modeling is aimed at combining diffusion model with another category of generative model to take all their advantages, which could provide stronger expressiveness or higher sampling speed. DiffuseVAE \cite{2201.00308} merges a standard VAE into the DDPM pipeline by conditioning diffusion sampling process with blurry image reconstructions generated by VAE. LSGM \cite{vahdat2021score} trains SGMs in the latent space of VAE, which generalizes SGMs into non-continuous data and enables smoother SGMs learning in a small space. Denoising diffusion GANs \cite{xiao2021tackling} introduces conditional GANs into DDPM pipeline to parameterize denoising process with a more expressive multimodal distribution, which provides large denoising steps. DiffFlow \cite{zhang2021diffusion} integrates flow function into trajectories of SDE-based diffusion model, which makes forward steps also trainable. The introduced randomness from noise perturbation endows normalizing flow with stronger expression power while the trainable forward process substantially reduce the diffusion trajectory length. Therefore, DiffFlow is able to learn distribution with sharper boundaries with better sampling efficiency.





\subsection{Multimodal Models}\label{sec:4.2}
Multimodal generation serves as an essential part in nowadays AIGC. 
The goal of multimodal generation is to learn a model that generates raw modalities by learning the multimodal connection and interaction from data~\cite{liang_foundations_2022}.
This connection and interaction between modalities can sometimes be very intricate, 
which makes the multimodal representation space hard to learn compared to the unimodal one.
However, with the emergence of the powerful modality-specific foundation architectures mentioned in previous sections, a growing number of methods are proposed in response to this challenge.
In this section, we introduce the state-of-the-art multimodal models in vision language generation, text audio generation, text graph generation and text code generation.
Since most multimodal generative models are always highly related to real-world applications, this section will mainly introduce from the perspective of downstream tasks.

\subsubsection{Vision Language Generation}
\begin{figure*}[t]
    \centering
    \includegraphics[width=0.9\linewidth]{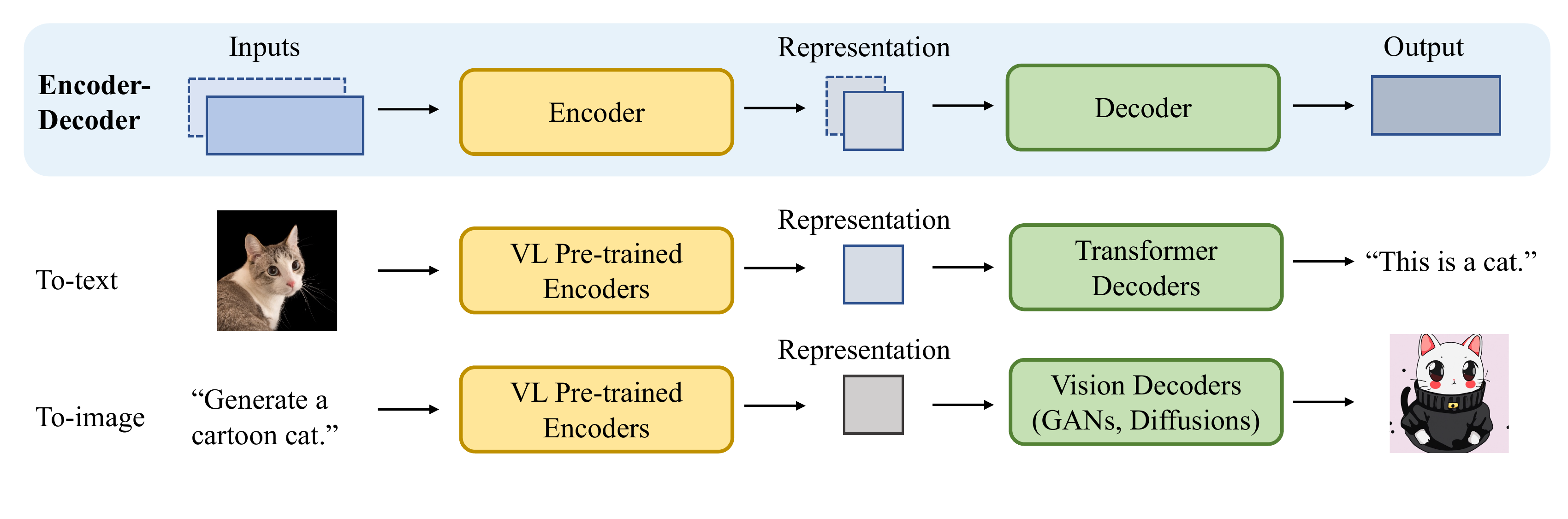}
    \caption{The general structure of generative vision language. 
    We separate the generation process into encoder part and decoder part. Encoder models will encode the inputs into a latent representation and then the decoder will decode this representation into a generated output.
    }
    \label{Fig:vl_general}
\end{figure*}

The encoder-decoder architecture is a widely used framework for solving unimodal generation problems in computer vision and natural language processing. In multimodal generation, particularly in vision-language generation, this method is often used as a foundation architecture. The encoder is responsible for learning a contextualized representation of the input data, while the decoder is used to generate raw modalities that reflect cross-modal interactions, structure, and coherence in the representation.
In the following, we present a comprehensive survey of state-of-the-art vision-language encoders, followed by an exposition of the decoder component.

\header{Vision Language Encoders}
\begin{figure*}[t]
    \centering
    \includegraphics[width=0.9\linewidth]{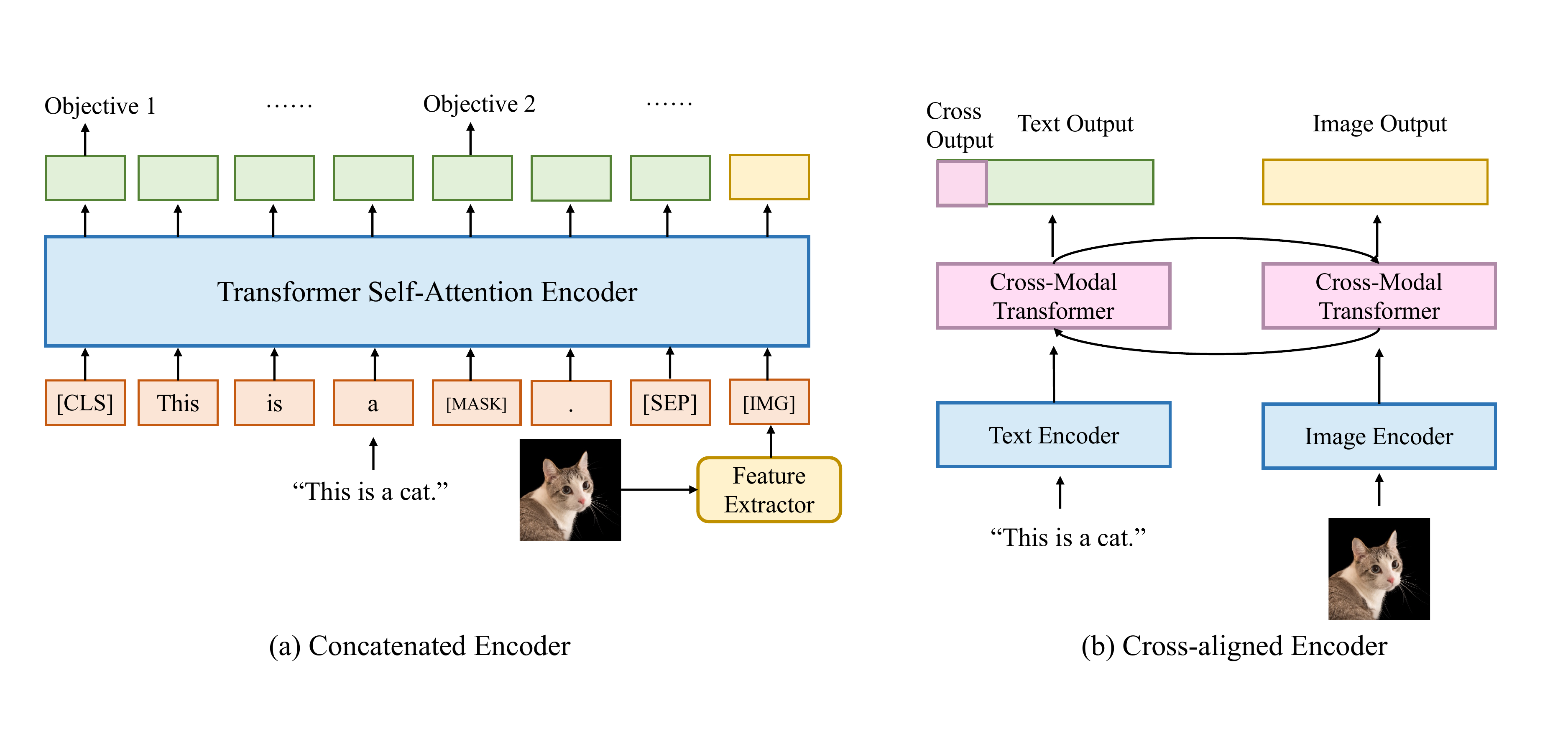}
    \caption{Two types of vision language encoders: concatenated encoders and cross-aligned encoders. Concatenated encoders accepts concatenated embeddings from different raw modalities, and cross-aligned encoders are aligned in abstract modalities.}
    \label{Fig:vl_encoder}
\end{figure*}
Recently, the development of encoders for single modalities has advanced significantly, leading to the question of how to learn contextualized representations from multiple modalities.
A common way to do this is to combine modality-specific encoders using a fusion function and then leverage multiple pre-training tasks to align the representation space~\cite{li2019visualbert, clip2021, wang2021simvlm}. Generally. these encoder models could be separated into two categories, concatenated encoders and cross-aligned encoders~\cite{liang_foundations_2022}. 

\textbf{Concatenated Encoders.}
A straight-forward solution to this problem is by concatenating the embeddings from single encoders.
An early example is VisualBERT~\cite{li2019visualbert}, which leverages BERT as text encoder, CNN as image encoder. The embeddings from the image encoder will be directly incorporated into BERT input embeddings, allowing the model to implicitly learn the aligned joint representation space. VisualBERT also leverages the multi-task pre-training paradigm as BERT, using two visually-grounded language model objectives: masked language modeling with image and sentence image prediction. Additionally, VisualBERT also incorporated some modality-specific pre-trianing objectives.
Another example is VL-BERT~\cite{su2019vl}, which shares the similar architecture as VisualBERT. Different from VisualBERT, VL-BERT uses Faster R-CNN~\cite{ren2015faster} as regions of interest (ROI) extractor, and leverages this extracted ROI information as the image region embedding. VL-BERT also includes an additional pre-training task, masked ROI classification with linguistic clues, for better incoporating the visual information.
Later, UNITER~\cite{zhou2020unified} was proposed based on the same architecture as VisualBERT, but with different training objectives. UNITER uses masked language modeling, masked region modeling, image text matching prediction and word region alignment prediction as its pre-training tasks. In this way, UNITER could learn informative contextualized embeddings.
To this end, we see that concatenated encoders are generally based on the same BERT architecture, and pre-trained with BERT-like tasks. 
However, these models always involves a very complicated pre-training process, data collection and loss design.
To solve this problem, \cite{wang2021simvlm} proposed SimVLM, which simplified the pre-training procedure of vision language models by setting PrefixLM as the training objective and directly using ViT as both text encoder and image encoder.
SimVLM achieved state-of-the-art performance on multiple vision language tasks compared with previous methods with a much simplified architecture.

\textbf{Cross-aligned Encoders.}
In addition to concatenating embeddings as input to encoders, another way to learn contextualized representations is to look at pairwise interactions between modalities~\cite{liang_foundations_2022}. 
Different from concatenated encoders, cross-aligned encoders always use a two-tower structure, where one tower for each modality and then learn a joint representation space using a cross-modality encoder.
LXMERT~\cite{tan2019lxmert} uses Transformers to extract image features and text features, and then adds a multimodal cross-attention module for coordination learning. The resulting output embeddings would be visual embeddings, language embeddings and multimodal embeddings.
The model is also pre-trained with several multimodal tasks.
Similarly, ViLBERT~\cite{vilbert} leverages a cross-transformer module to align the two modalities. Given vision and language embeddings, the keys and values of one certain modality will be input into another modality's attention module to generate a pooled attention embedding that incorporates both information. 
In general, these models all leverage a cross layer to fuse the information into a joint representation space. Nevertheless, employing a transformer architecture in this context would be inefficient due to its large number of parameters.
To simplify the training process and calculation, CLIP~\cite{clip2021} uses dot product as the cross layer, which is more efficient than the transformer encoder, enabling efficient large-scale downstream training. Furthermore, CLIP is trained on copious amounts of pairwise data, which has been shown to outperform numerous other models.

\header{Vision Language Decoders}
\begin{figure*}[t]
    \centering
    \includegraphics[width=0.9\linewidth]{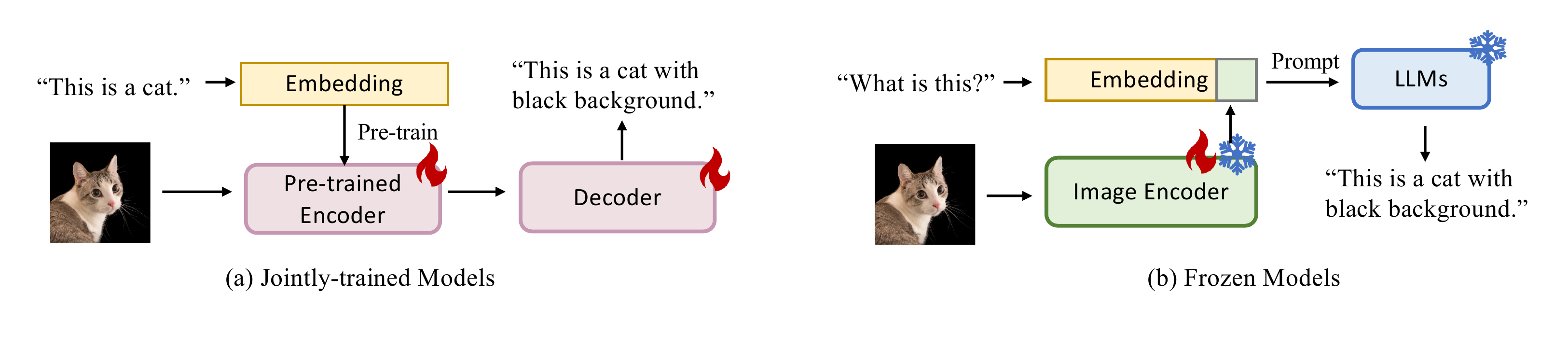}
    \caption{Two types of to-language decoder models: jointly-trained models and frozen models. Jointly-trained models are normally trained end-to-end, while frozen models normally keep the language decoder frozen and only train the image encoder. }
    \label{Fig:vl_decoder}
\end{figure*}
Given a representation from a certain modality, vision language decoder mainly aims to transform it into a certain raw modality as specified by the task.
In this section, we will mainly focus on to-text and to-image decoders.

\textbf{To-text decoders.}
To-text decoders generally take in contextualized representations from the encoder and decode the representation into a sentence.
With the emergence and proven effectiveness of large language models, many architectures are now selectively freezing the language decoder component. As a result, to-text decoders can be broadly categorized into two types: jointly-trained models and frozen models.

\textit{Jointly-trained decoders.}
Jointly-trained decoders refer to decoders that require complete cross-modal training when decoding representation. 
The challenge of text-to-text generation typically lies in aligning the two modalities during pre-training. As a result, the model requires a stronger encoder rather than a decoder. To address this challenge, many models prioritize constructing a strong encoder and then combine it with a relatively lightweight decoder model. For example, VLP~\cite{zhou2020unified} and ALBEF~\cite{li2021align} leverage a simple transformer decoder to decode the information.
BLIP~\cite{li2022blip} combines an encoder and decoder together during pre-training, allowing for multimodal space alignment for both understanding and generation objectives. BLIP is composed of three parts, a unimodal encoder for extracting image and text features, an image-grounded text encoder which accepts image and text features as input, and an image-grounded text decoder, which accepts image features and outputs text. Except for the aligned encoder and decoder structure, the authors also designed several corresponding pre-training tasks to help the model better learn the multimodal dependency.

\textit{Frozen deocders.}
Another way to efficiently perform to-text generation tasks is to freeze the large language model and train an image encoder only, which can also be seen as a way to perform multimodal prompting.
Due to the success of prompting and in-context learning in NLP, there has been increased attention towards methods of this nature. This has led people to question whether such methods could be effective in multimodal settings as well.
Frozen~\cite{tsimpoukelli2021multimodal} first introduced in-context learning into vision language tasks. It freezes the language model and only trains the image encoder. The produced image representations will be embeded in the input embeddings of the language model. This method achieves state-of-the-art performance in various zero-shot and few-shot vision language tasks.
Later, Alayrac et al. proposed Flamingo~\cite{alayrac2022flamingo}, which further explored multimodal in-context learning.
Flamingo involves a frozen vision encoder and a frozen language encoder to get vision language representations, and utilizes gated cross-attention-dense layer to fuse the image representation into text representation.
Recently, \cite{koh2023grounding} proposed a method to realize VL dialogue with frozen language models, enabling the model to generate interleaved multimodal data.
This method also freezes input encoders and train text-to-image and image-to-text linear maps to further encode and decode produced embeddings.
However, it still remains a question why this kind of prompting based method work in multimodal generation. Some works have also been proposed to answer this question. Merullo et al. proposed a method~\cite{merullo2022linearly} that injects a linear projection between the frozen image encoder and the text encoder. During training, only the linear projection is tuned. The experiment results show that frozen language models with similar sizes generally perform equally well at transferring visual information into language, but image encoders pre-trained with linguistic supervision like CLIP text encoder, could encode extra information and thus perform significantly better on vision language tasks.
%

\textbf{To-image decoders.}
To-image generation refers to given an instruction, generating an image that corresponds to the instruction. Similarly, 
commonly used models in image generation also follow an encoder-decoder architecture, where the encoders are more focused on learning language information and the decoders are more focused on leveraging the learned information to restrict image synthesis.
Generally, recent works could be separated into two categories, GAN-based methods and diffusion-based methods.

\textit{GAN-based decoders.}
Given a text encoder $\phi(t)$, GAN-based methods combine a discriminator $D$ and a generator $G$, where the generator $G$ accepts the text embedding generated by $\phi(t)$ and noise vector $z$ to generate output $X_g$, which are input to the discriminator $D$ with the real sample distribution $X_r$~\cite{zhou2021survey}.
A notable model in this area is StackGAN~\cite{zhang2017stackgan}. StackGAN architecture consists of two stages: a conditioning stage and a refinement stage. In the conditioning stage, the model takes in the textual description as input and generates a low-resolution image. This image is then fed into the refinement stage, where it is further refined to produce a high-resolution image that matches the textual description.
AttnGAN~\cite{xu2018attngan} is another text-to-image synthesis model that builds upon the StackGAN architecture. Attngan adds an attention mechanism to the StackGAN architecture to further improve the quality of generated images. 
However, these models mainly uses a comparatively simple text encoder during instruction learning, which could lead to certain information loss.
StyleCLIP~\cite{patashnik2021styleclip} is a recent model for text-to-image synthesis that uses contrastive learning to align text and image features. It is based on the StyleGAN~\cite{karras2019style} architecture and represents a significant advancement over previous text-to-image synthesis models such as StackGAN. StyleCLIP also follows the encoder-decoder structure that use a text encoder to encode instructions and an image decoder to synthesize a new image. One of the key innovations of StyleCLIP is its use of contrastive learning to align the text and image features. By training the model to maximize the similarity between the text and image features while minimizing the similarity between different text and image pairs, StyleCLIP is able to learn a more effective mapping between text and image features, resulting in higher-quality image synthesis.


\textit{Diffusion-based decoders.}
\begin{figure*}[t]
    \centering
    \includegraphics[width=0.8\linewidth]{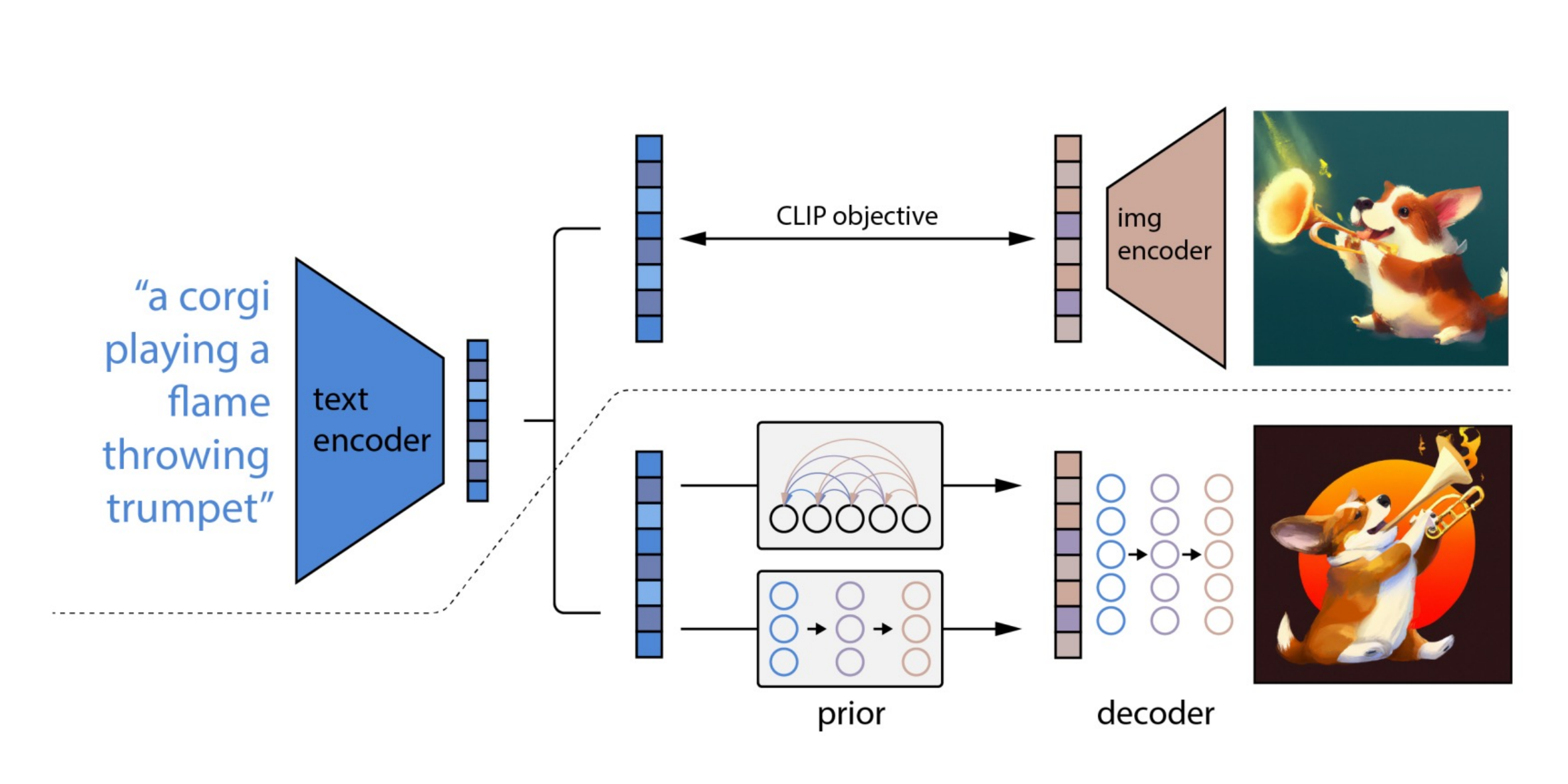}
    \caption{The model structure of DALL-E-2. Above the dotted line is the CLIP pre-training process, which aims to align the vision and language modalities. And below the dotted line is the image generation process. The text encoder accepts an instruction and encodes it into a representation, then the prior network and diffusion model decodes this representation to generate the final output. }
    \label{Fig:vl_dalle2}
\end{figure*}
Generative image modelling has recently seen great success with the use of diffusion models. These models have also been applied in text-to-image generation.
For example, GLIDE~\cite{glide} introduces ablated diffusion model (ADM) into text-to-image generation. Compared to previous diffusion based methods, GLIDE uses larger model with 3.5B parameters and larger pairwise datasets, which achieved better results on many benchmarks. 
Different from GLIDE, Imagen~\cite{saharia2022photorealistic} combines a frozen T5 language model with a super-resolution diffusion model. The frozen encoder will encode the text instruction and generates an embedding, then the first diffusion model will accordingly generate an low-resolution image. The second diffusion model accepts this image with the text embedding and outputs a high-resolution image.
DALL-E-2~\cite{ramesh_hierarchical_2022} combines CLIP encoder with diffusion decoder for image genration and editing tasks. Compared with Imagen, DALL-E-2 leverages a prior network to translation between text embedding and image embedding.
Except for advancement in model design, another major difference between these diffusion based models and previous generative methods is that these diffusion based models are commonly trained on larger dataset with much more parameters, which make them possible to learn better representations over others.

In addition to previously mentioned methods, there are also works that use VAE as the decoder. For example, Ramesh et al. proposed DALL-E~\cite{dalle}, a zero-shot image generator that utilizes dVAE as image encoder and decoder, BPE as text encoder and pre-trained CLIP during inference. 


\subsubsection{Text Audio Generation}

The field of text-audio multimodal processing has seen significant growth in recent years. Most models in this field focus on either synthesis tasks, such as speech synthesis, or recognition tasks, such as automatic speech recognition. They refer to the process of converting written text into spoken speech or accurately transcribing human speech into machine-readable text. However, text audio generation is a distinct task that involves creating novel audio or text using multimodal models. While related, text-audio generation, synthesis, and recognition tasks differ in their goals and the techniques used to achieve them. In this work, we focus on text-audio generation rather than synthesis or recognition tasks.

\paragraph{Text-Audio Generation.}
AdaSpeech~\cite{chen2021adaspeech} is proposed to efficiently customize new voices with high quality using limited speech data by utilizing two acoustic encoders and conditional layer normalization in the mel-spectrogram decoder. Since previous studies have limitations in style conversion, Lombard~\cite{paul2020enhancing} exploits the Spectral Shaping and Dynamic Range Compression~\cite{zorila2012speech} to generate highly intelligible speech in the presence of noise. Cross-lingual generation is another Influential work to transfer voices across languages. \cite{zhang2019learning} can produce high-quality speech in multiple languages and transfer voices across languages through the use of phonemic input representation and adversarial loss term to disentangle speaker identity from speech content.

\paragraph{Text-Music Generation.}
\cite{yu2019deep} proposes a deep cross-modal correlation learning architecture for audio and lyrics, where intermodal canonical correlation analysis is used to calculate the similarity of temporal structures between audio and lyrics. To better learn social media content, JTAV~\cite{liang2018jtav} fuses textual, acoustic, and visual information using cross-modal fusion and attentive pooling techniques. Different from JTAV, \cite{ferraro2021enriched} combines multiple types of information more related to music, such as playlists-track interactions and genre metadata, and align their latent representations to model unique music piece. In addition, there are some works focusing on generating text information, such as descriptions and captions, given the audio as input. \cite{choi2016towards} is proposed to generate descriptions for music playlists by combining audio content analysis and natural language processing to utilize the information of each track. MusCaps~\cite{manco2021muscaps} is a music audio captioning model that generates descriptions of music audio content by processing audio-text inputs through a multimodal encoder and leveraging audio data pre-training to obtain effective musical feature representations. For music and language pre-training, Manco et al.~\cite{manco2022learning} propose a multimodal architecture, which uses weakly aligned text as the only supervisory signal to learn general-purpose music audio representations. CLAP \cite{elizalde2022clap} is another method for learning audio concepts from natural language supervision that utilizes two encoders and contrastive learning to bring audio and text descriptions into a joint multimodal space.

\subsubsection{Text Graph Generation}

Text-graph generation is an essential multi-modal topic which could largely free the potential of NLP systems. Natural language text is intrinsically vague as it carries various redundant information and is also weakly organized in logic. Meanwhile, it is favorable for machines to work with structured, well-organized and compressed form of contents. Knowledge graph (KG) is structural meaning representation which reflects relationships among semantic internal states as graph structure in a language processing system. And there are increasing number of works extracting KG from text to assist text generation which incorporates complicated ideas across multiple sentences. Semantic parsing can also be formulated into a problem of text-to-graph generation. It aims to convert natural language text to a logical form, mostly abstract meaning representation (AMR) \cite{banarescu2013abstract}, which is a broad-coverage sentence-level semantic representation. Compared to text-to-KG generation, it emphasizes on providing machine interpretable representations rather than constructing a semantic network. Conversely, KG-to-text generation aims to generate fluent and logically-coherent text based on already constructed KG. Apart from the domain of NLP, text-graph generation could also push forward the boundary of computer aided drug design. There are emerging works bridging highly structured molecule graph with language descriptions, which facilitates human understanding of profound molecular knowledge and novel molecule exploration. In the following, we briefly overview some representative works in these four topics.

\paragraph{Text To Knowledge Graph Generation.}
Li et al. \cite{li2016commonsense} treat text-to-KG construction as a process of knowledge graph completion (KGC), where missing terms are progressively covered by inference. A bilinear model and another DNN-based model are adopted to embed terms and compute score of arbitrary tuples for additive operation. KG-BERT \cite{yao2019kg} utilizes the power of pre-trained language models to capture more contextualized information during KGC. The idea is to represent triplets as textual sequences and models graph completion as a sequence classification problem by fine-tuned BERT model. Malaviya et al. \cite{malaviya2020commonsense} propose an approach incorporating graph convolutional network (GCN) for to extract more structural and semantic context. It also tackles graph sparsity and scalability issues by introducing graph augmentation and progressive masking strategies. Alternatively, another line of works \cite{petroni2019language,shin2020autoprompt,li2021prefix} directly query pre-trained language models to obtain a semantic knowledge network. Specifically, language models are repeatedly prompted to predict the masked terms in cloze sentence to acquire relational knowledge. CycleGT \cite{2006.04702} is an unsupervised method allowing text-KG translation in both directions. An unsupervised cycle training strategy is adopted to provide self-supervision, which enables the entire training process possible with non-parallel text and graph data. Utilizing similar strategy, DualTKB \cite{2010.14660} further proves that model performance could be largely improved even under a weakly supervised setting. Lu et al. \cite{lu2022unified} propose a unified text-to-graph framework which incorporates most information extraction tasks. Meanwhile, the use of a pre-defined schema may limit its generalization to diverse text forms of nodes and edges. Grapher \cite{melnyk2022knowledge} performs end-to-end text-to-KG construction efficiently by generating node and edge in two separate stages. Specifically, a pre-trained language model is first fine-tuned with entity extraction tasks for node generation. Subsequently, focal loss and sparse adjacency matrix are introduced to address the skewed edge distribution issue during edge construction. 
\begin{figure*}[htp]
    \centering
    \includegraphics[width=0.9\linewidth]{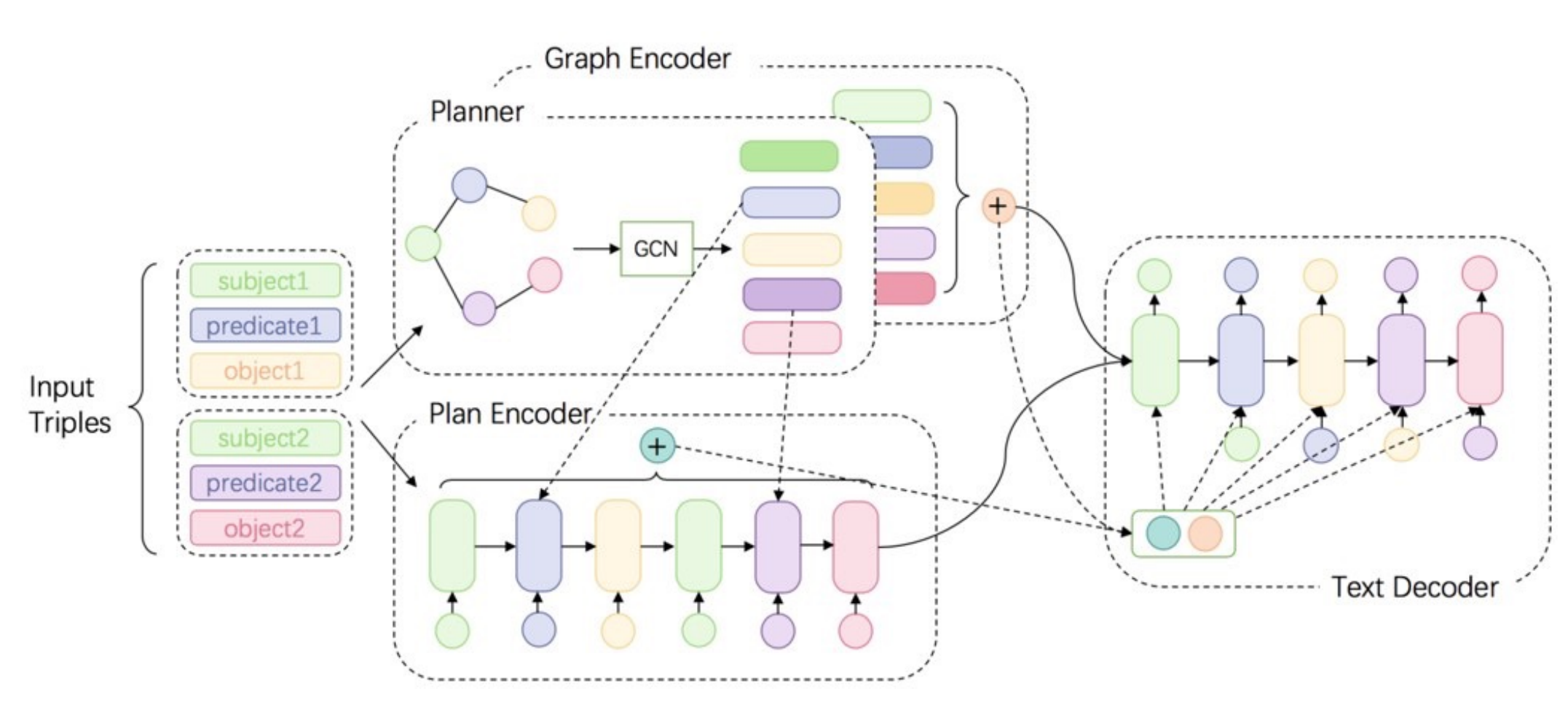}
    \caption{DUALENC \cite{zhao-etal-2020-bridging}: a KG-to-text generation model that bridges the structural gap between KG and graph via dual-encoding.}
    \label{Fig:dualenc}
\end{figure*}
\paragraph{Knowledge Graph To Text Generation.}
GTR-LSTM \cite{distiawan2018gtr} is a sequence-to-sequence encoder-decoder framework which generates text from linearized KG triples. It could handle cycles in KGs for capturing global information. Meanwhile, its linearized graph nature could still result in considerable structural information loss, especially for large graphs. To address this issue, Song et al. \cite{1805.02473} encode graph semantics with a graph-state LSTM which enables information propagation between nodes during a series of state transitions. It proves to be capable of modeling non-local interactions between nodes while also efficient due to high parallelization. Zhao et al. \cite{zhao-etal-2020-bridging} propose DUALENC, a dual encoding model, to bridge the structural discrepancy between input graph and output text. Specifically, it utilizes a GCN-based graph encoder to extract structural information, while a neural planner is also adopted to create a sequential content plan of a graph for generating linear output text. Alternatively, Koncel-Kedziorski et al. \cite{koncel2019text} encode graph structure for text generation with a transformer-based architecture extended from the graph attention network (GAT) \cite{velivckovic2017graph}. The idea is to compute the node representations of KG by traversing its local neighborhood with self-attention mechanism. In contrast, Ribeiro et al. \cite{ribeiro2020modeling} focus on utilizing local and global node encoding strategies jointly to capture complementary information from graph contexts. Adapted from transformer, HetGT \cite{yao2020heterogeneous} aims at modeling different relationships in the graph independently to avoid information loss by simply mixing them. The input graph is first transformed into a heterogeneous Levi graph and then split into sub-graphs based on the heterogeneity of each part for future information aggregation.  

\paragraph{Semantic Parsing.}
Early works \cite{dong2016language,jia2016data} formulate semantic parsing as sequence-to-sequence generation problems. However, AMR is a structured object by its nature. Sequence-to-sequence problem setup could only capture shallow word sequence information meanwhile potentially ignore abundant syntax and semantic information. Lyu et al. \cite{lyu2018amr} model semantic parsing as a graph prediction problem by expressing AMR as a root labeled directed acyclic graph (DAG) This would require an alignment between node in the graph and word in the sentences. A neural parser which treat alignments as a latent variable in a joint probabilistic model is proposed for node alignment and edge prediction during AMR parsing. Chen et al. \cite{chen2018sequence} construct semantic graph with an action set via a neural sequence-to-action RNN model. Parsing process are reinforced by integrating both structural and semantic constraints during decoding. Zhang et al. \cite{1905.08704} tackle issues emerged from the reentrancy property in AMR parsing via an aligner-free attention based model which formulate the problem into sequence-to-graph transduction. Utilizing a pointer-generator network, it is proved that the model can be trained effectively with limited labeled AMR data. Fancellu et al. \cite{fancellu2019semantic} propose a graph-aware sequential model to construct linearized graph for AMR graph prediction. Without a latent variable, it ensures each well-formed string will be only paired with exactly only one derivation by a novel graph-aware string rewriting strategy.
\begin{figure*}[htp]
    \centering
    \includegraphics[width=0.8\linewidth]{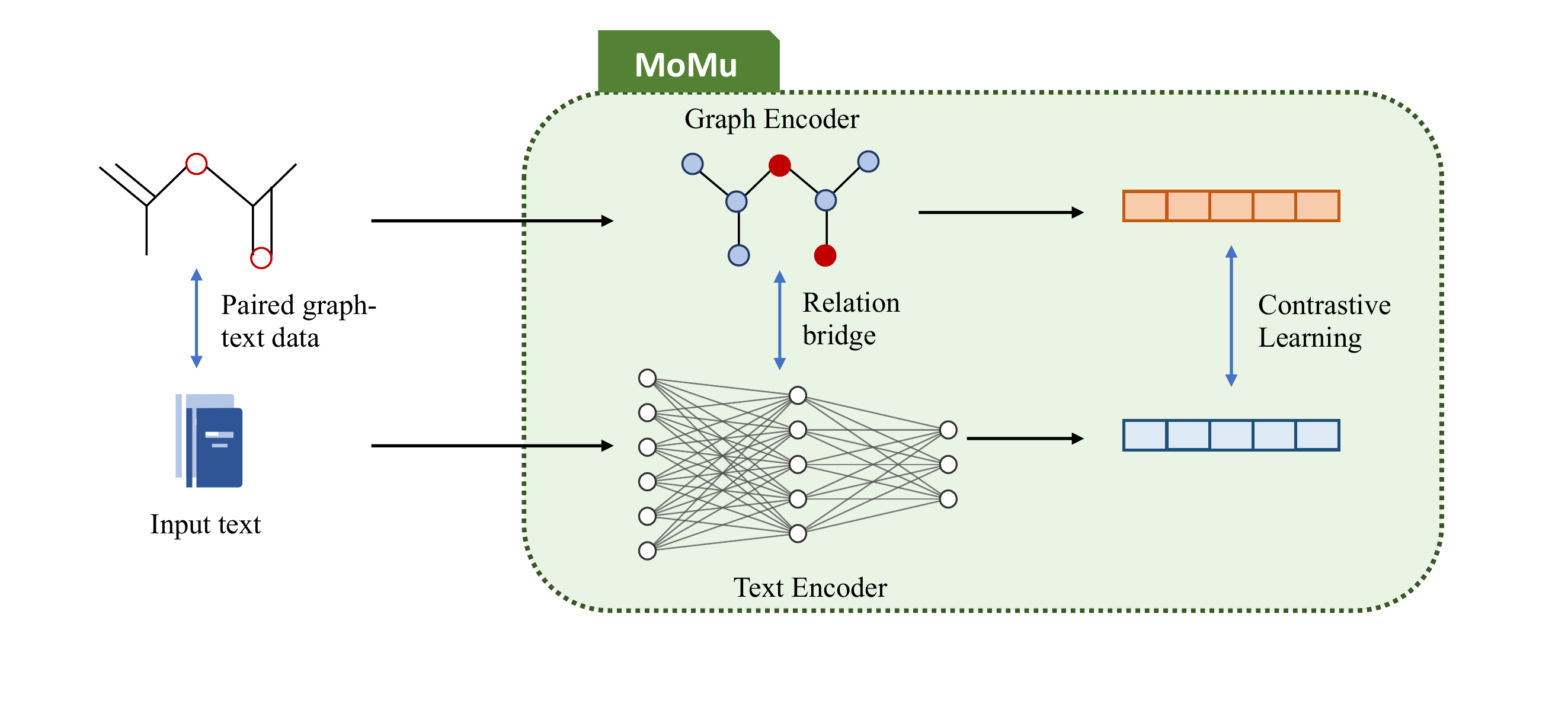}
    \caption{MoMu \cite{2209.05481}: A cross-modal text-molecule generation model.}
    \label{Fig:text_molecule}
\end{figure*}
\paragraph{Text Molecule Generation.}
Text2Mol \cite{inproceedings} is a cross-modal information retrieval system to retrieve molecule graph based on language description. A BERT-based text encoder and a MLP-GCN combined molecule encoder are utilized to create multi-modal embedding in a semantic space, which is aligned by contrast learning with paired data. Instead of retrieving from existing molecules, MolT5 \cite{2204.11817} proposes a self-supervised learning framework for text-conditioned de-novo molecule generation and molecule captioning. It tackles the scarcity of cross-modal data pair with a pre-train and fine-tune strategy. Specifically, it pre-trains the model on unpaired text and molecule strings with a denoising objective, followed by fine-tuning with limited paired data. However, restricted by its linearized graph nature, string-based representation of a molecule is not unique and could result in structural information loss. To tackle this issue, MoMu \cite{2209.05481} introduces a graph based multi-modal framework which trains two separate encoders jointly by contrast learning for semantic space alignment with weakly-paired cross-modal data. It can also be adapted to various downstream tasks apart from de-novo molecule graph generation.

\subsubsection{Text Code Generation}

Text-code generation aims to automatically generate valid programming code from natural language description or provide coding assist. LLMs have recently exhibited great potential in programming language (PL) code generation from natural language (NL) descriptions. Early works directly formulate text-code generation as a pure language generation task. However, NL and PL are data types with inherently different modalities, additional strategies are essential in capturing mutual dependencies between NL and PL during semantic space alignment. Compared to NL data, PL data also encapsulates rich structural information and different syntax, which makes it more challenging to understand semantic information from the PL context. Furthermore, text-code models are also expected to be multi-lingual as they could provide better generalization. In the following, we mainly introduce code generation models conditioned on NL description. We also review other coding assist models based on language. 

\paragraph{Text-conditioned Programming Code Generation.} 
CodeBERT \cite{feng2020codebert} is a bimodal Transformer-based pre-trained text-code model which could capture the semantic connection between NL and PL. It adopts a hybrid objective function by utilizing binomial NL-PL paired data for model training and unimodal PL code data for learning better generators respectively to align NL and PL in semantic space. This model is further pre-trained on six multi-lingual PL for better generalization. CuBERT \cite{kanade2020learning} shares similar model architecture with CodeBERT meanwhile it is not required to perform sentence separation between the natural-language description of a function and its body for sentence-pair representation. CodeT5 \cite{wang2021codet5} proposes a pre-trained encoder-decoder Transformer model which better captures contextualized semantic information from code. Specifically, it introduces novel identifier-aware pre-training tasks to preserve crucial token type information by discriminating identifiers from code tokens and recover them when masked. PLBART \cite{ahmad2021unified} extends bimodal text-code model from generative tasks to a broader categories of discriminative tasks such as clone and vulnerable code detection under a unified framework. Another line of works \cite{yin2017syntactic,dai2018syntax} introduce the notion of program graphs \cite{allamanis2017learning} to explicitly model the structures underlying PL code to assist generation. The program graphs are constructed as Abstract Syntax Trees (AST) to encapsulate knowledge from program-specific semantic and syntax.

\paragraph{Interactive Programming System.}
Text-code generation are jointly challenged by the intractable searching space of programming code generation and improper specification of user intent due to the intrinsic ambiguity of NL. CODEGEN \cite{2203.13474} propose a multi-turn program synthesis approach which factorizes program synthesis conditioned on a single complicated NL specification into progressive generation controlled by a series of user intents. It is constructed in the form of autoregressive transformers learning a conditional distribution of the next token given previous tokens and it is trained on both PL and NL data. TDUIF \cite{2208.05950} extends interactive programming framework by formalizing the user intent and providing more understandable user feedback. It further realizes scalable automatic algorithm evaluation which does not require user in loop with high-fidelity user interaction modeling.

\section{Applications}\label{sec:5}
\begin{figure*}[htp]
    \centering
    \includegraphics[width=0.8\linewidth]{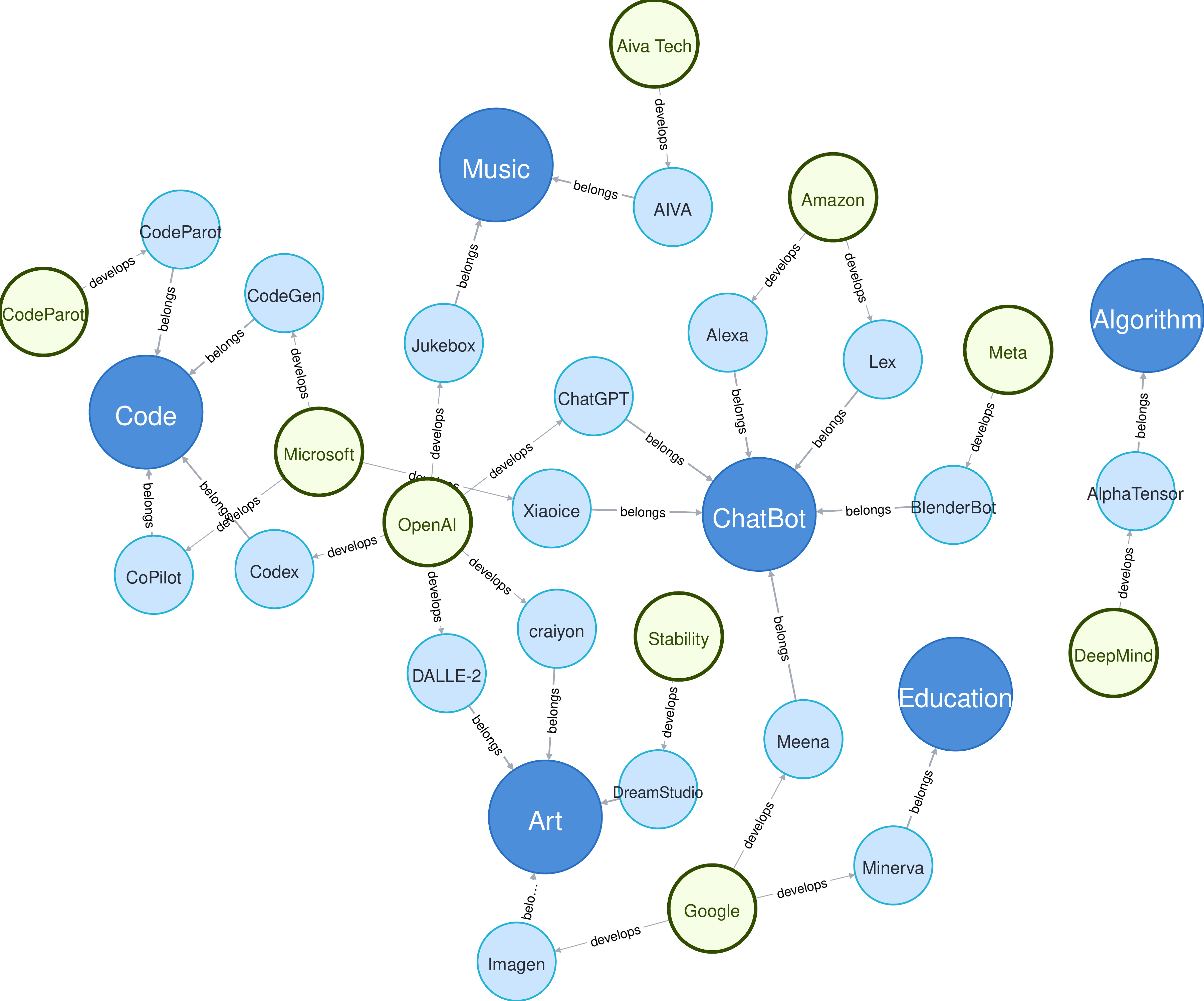}
    \caption{A relation graph of a current research areas, applications and related companies, where dark blue circles represent research areas, light blue circles represent applications and green circles represents companies.}
    \label{Fig:text_molecule}
\end{figure*}

\begin{table}[!htbp]
    \centering
    \resizebox{0.9\textwidth}{!}{
    \begin{tabular}{llllll}
    \hline
    \textbf{Application} & \textbf{Platform/Software} & \textbf{Company} & \textbf{Year} & \textbf{Papaer} & \textbf{Link} \\ \hline
ChatBot             & Xiaoice     & Microsoft  & 2018 & \cite{zhou2020design}               & \href{https://www.xiaoice.com/}{Xiaoice}                                                           \\
ChatBot             & Meena       & Google     & 2020 & \cite{adiwardana2020towards}        & \href{https://ai.googleblog.com/2020/01/towards-conversational-agent-that-can.html}{Meena Blog}    \\
ChatBot             & BlenderBot  & Meta       & 2022 & \cite{shuster2022blenderbot}        & \href{https://blenderbot.ai/}{Blenderbot}                                                          \\
ChatBot             & ChatGPT     & OpenAI     & 2022 & \cite{ouyang_training_2022}         & \href{https://chat.openai.com/chat}{ChatGPT}                                                       \\
ChatBot             & Alexa       & Amazon     & 2014 & -                                   & \href{https://alexa.amazon.com/}{Amazon Alexa}                                                     \\
ChatBot             & Lex         & Amazon     & 2017 & -                                   & \href{https://aws.amazon.com/lex/}{Amazon Lex}                                                     \\
Music               & AIVA        & Aiva Tech  & 2016 & -                                   & \href{http://www.aiva.ai}{AIVA}                                                                    \\
Music               & Jukebox     & OpenAI     & 2020 & \cite{dhariwal2020jukebox}          & \href{https://openai.com/blog/jukebox/}{Jukebox}                                                   \\
Code                & CodeGPT     & Microsoft  & 2021 & \cite{lu2021codexglue}              & \href{https://github.com/microsoft/CodeXGLUE}{CodeGPT}                                             \\
Code                & CodeParrot  & CodeParrot & 2022 & \cite{tunstall2022natural}          & \href{https://huggingface.co/codeparrot/codeparrot}{CodeParrot}                                    \\
Code                & Codex       & OpenAI     & 2021 & \cite{chen_evaluating_2021}         & \href{https://openai.com/blog/openai-codex/}{Codex blog}                                           \\
Code                & CoPilot     & Microsoft  & 2021 & \cite{chen_evaluating_2021}         & \href{https://github.com/features/copilot}{CoPilot}                                                \\
Art                 & DALL-E-2    & OpenAI     & 2022 & \cite{ramesh_hierarchical_2022}     & \href{https://openai.com/dall-e-2/}{DALL-E-2 Blog}                                                 \\
Art                 & DreamStudio & Stability  & 2022 & \cite{rombach_high-resolution_2022} & \href{https://beta.dreamstudio.ai/home}{Dreamstudio}                                               \\
Art                 & craiyon     & OpenAI     & 2021 & \cite{ramesh2021zero}               & \href{https://www.craiyon.com/}{Craiyon}                                                           \\
Art                 & Imagen      & Google     & 2022 & \cite{saharia2022photorealistic}    & \href{https://imagen-ai.com/?v=1c}{Imagen}                                                        
      \\
Education           & Minerva     & Google     & 2022 & \cite{lewkowycz2022solving}         & \href{https://ai.googleblog.com/2022/06/minerva-solving-quantitative-reasoning.html}{Minerva Blog} \\
Algorithm & AlphaTensor & DeepMind   & 2022 & \cite{fawzi2022discovering}         & \href{https://github.com/deepmind/alphatensor}{AlphaTensor}                                        \\
\hline
    \end{tabular}}
    \caption{Applications of Generative AI models.}
    \label{tab:applications}
\end{table}

\subsection{ChatBot}
A chatbot is a computer program designed to simulate conversation with human users through text-based interfaces. 
Chatbots normally use language models to understand and respond to user queries and inpus in a conversational manner.
They can be programmed to perform a wide range of tasks, for example, providing customer support and answering frequently asked questions.
One of the most prominent example is Xiaoice~\cite{zhou2020design}. 
XiaoIce was developed by a team of researchers and engineers from Microsoft, using state-of-the-art techniques in natural language processing, machine learning, and knowledge representation.
An important feature of Xiaoice is that it is able to express empathy, which is achieved by using sentiment analysis methods, to make Xiaoice perform like a human.
In 2020, Google proposed Meena~\cite{adiwardana2020towards}, a multi-turn open-domain
chatbot trained on social media conversations, which achieves state-of-the-art interactive SSA score and perplexity.
Recently, Microsoft released their newest version Bing, which incorporates ChatGPT, enabling its users to ask open domain or conditioned questions and get results through conversation. 
This presents new possibilities for the development of chatbots in the future.

\subsection{Art}
AI art generation refers to using computer algorithms to create original works of art. These algorithms are trained on large datasets of existing artwork and use machine learning techniques to generate new pieces that mimic the styles and techniques of famous artists or explore new artistic styles. 
With the rapid development in diffusion based models, more and more companies have launched their art generation products.
One of the most notable advancements in the field is the DALL-E series, which was introduced by OpenAI. DALL-E~\cite{ramesh2021zero}, which is now Craiyon, was first built on VQ-VAE and CLIP, then diffusion was also applied to this product, becoming DALL-E-2~\cite{ramesh_hierarchical_2022}.
DreamStudio~\cite{rombach_high-resolution_2022}, created by Stability.ai, is a text-to-image generation service that utilizes stable diffusion to generate images based on given phrases or sentences. This technology offers comparable performance to that of DALL-E-2, but with even faster processing speeds, making it a popular choice for many users.
Imagen~\cite{saharia2022photorealistic}, developed by Google, uses diffusion in its image editing and generation service. In a blog post, Google reported that they conducted a study with human raters to evaluate the quality of AI-generated images. The results showed that Imagen outperformed other models in side-by-side comparisons, with higher ratings for sample quality and image-text alignment preferred by the human raters.

\subsection{Music}
Deep music generation refers to the use of deep learning techniques and artificial intelligence algorithms to generate novel and original pieces of music. A prominent approach is to produce a symbolic representation of the music in the form of a piano roll. This approach entails specifying the timing, pitch, velocity, and instrument for each note to be played. AIVA~\footnote{http://www.aiva.ai} is one of the most notable examples, which is developed by Aiva Technologies in 2016. It can generate music clips in multiple styles including electronic, pop, jazz, etc. and can be used in various contexts. As the world's first artificial intelligence composer recognized by symphonic organizations, AIVA obtained the global status of Composer in the SACEM music society. OpenAI develops Jukebox~\cite{dhariwal2020jukebox} in 2020. It generates music with singing in the raw audio domain in diverse genres and artistic styles. Jukebox is considered as a leap forward in terms of musical quality, coherence, audio sample duration, and the capacity to be conditioned by artist, genre, and lyrics.

\subsection{Code}
AI-based programming systems generally aim for tasks including code completion, source code to pseudo-code mapping, program repair, API sequence prediction, user feedback, and natural language to code generation. Recently, the emergence of powerful LLMs has pushed the boundary of AI-based programming a large step forward. CodeGPT \cite{lu2021codexglue} is an open-source code generation model developed by OpenAI which follows the transformer architecture as many other models in the GPT family. It can be fine-tuned for various code generation tasks such as code completion, summary, or translation based on a vast amount of source code data. CodeParrot \cite{tunstall2022natural} is a programming learning platform that provides user with personalized feedback and assistance during coding. A variety of interactive exercises and programming challenges are designed in the fashion of progressive human-machine interaction. One unique feature is the scaffolding strategy which splits complicated tasks into smaller and manageable steps to help students gradually build their coding skills. Trained on a much larger and more diverse corpus of data, Codex \cite{chen_evaluating_2021} is a significant step forward compared to most previous models. Specifically, it is designed to generate complete coding programs from scratch while CodeGPT is only able to generate code fragments that complete a given prompt. It also enjoys the benefits of being adapted to multiple programming languages, which could provide better flexibility and generalization.

\subsection{Education}
AIGC has the potential to achieve significant advancements in education by leveraging multi-modality data, for example, tutorial videos, academic papers, and other high-quality information, thereby improving the personalized education experience. On the academic side, Google Research introduced Minerva~\cite{lewkowycz2022solving}, which is built upon PaLM general language models~\cite{chowdhery2022palm} and an additional science-and-math-focused dataset, to solve college-level multi-step quantitative tasks, covering algebra, probability, physics, number theory, precalculus, geometry, biology, electric engineering, chemistry, astronomy, and machine learning. For example, it can give step-by-step details of proving the inequality $a^2+b^2\geq 2ab$ for any $(a,b)\in\mathbb{R}^2$ and it can also correctly identify Americium as the radioactive element among other three choices, including Sodium, Chromium, and Aluminum.
As is described in the blog\footnote{https://ai.googleblog.com/2022/06/minerva-solving-quantitative-reasoning.html}, Minerva achieves state-of-the-art performance on reasoning tasks by combing techniques, including few-shot prompting, a chain of thought or scratchpad prompting, and majority voting. Although Minerva's performance is still below human performance, with continuous improvement and future advancement, AIGC could provide affordable personalized math tutors. On the commercial side, Skillful Craftsman Education Technology announced to develop a class bot product powered by AIGC and featuring auto curriculum, AI tutor, and self-adaptive learning for online education, which is expected to be shipped by the fourth quarter of 2023.








\section{Efficiency in AIGC}\label{sec:6}

Deep generative AI models with neural networks has dominated the field of machine learning for the past decade, with its rise attributed to the ImageNet competition in 2012~\cite{deng2009imagenet}, which led to a race to create deeper and more complex models. This trend is also seen in natural language understanding, where models like BERT and GPT-3 have been developed with a large number of parameters. However, the increasing model footprint and complexity, as well as the cost and resources required for training and deployment, pose challenges for practical deployment in the real world. The core challenge is efficiency, which can be broken it down as follows:
\begin{itemize}
    \item \textbf{Inference efficiency}: This is concerned with the practical considerations of deploying a model for inference, i.e., computing the model's outputs for a given input. Inference efficiency is mostly related to the model's size, speed, and resource consumption (e.g., disk and RAM usage) during inference.
    \item \textbf{Training efficiency}: This covers factors that affect the speed and resource requirements of training a model, such as training time, memory footprint, and scalability across multiple devices. It may also encompass considerations around the amount of data required to achieve optimal performance on a given task.
\end{itemize}

\subsection{Prompt Learning}
\begin{figure*}[t]
    \centering
    \includegraphics[width=0.9\linewidth]{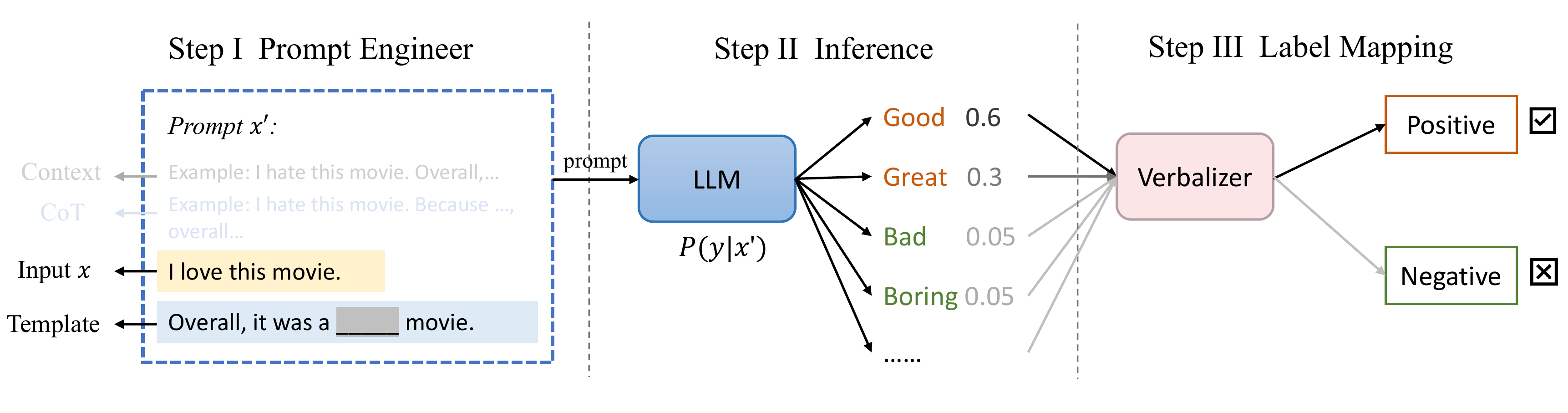}
    \caption{General procedure of prompt learning for emotion detection examples. First, the user need to construct a prompt that fits the problem well, the user can also use in-context learning and chain-of-thought (CoT) to help improve the performance. Then, an LLM will generate suitable words for the blank space in the prompt. Finally, a verbalizer will project the generated word to a specific classification category.}
    \label{Fig:prompt}
\end{figure*}

Prompt learning is a relatively new concept that has been proposed in recent years within the context of pre-trained large language models. Previously, to make a prediction $y$ given input $x$, the goal of traditional supervised learning is to find a language model that predicts 
 the probability $P(y|x)$. With prompt learning, the goal becomes finding a template $x'$ that directly predicts the  probability $P(y|x')$~\cite{liu2023pre}. 
 Hence, the objective of using a language model becomes encouraging a pre-trained model to make predictions by providing a prompt specifying the task to be done.
 Normally, prompt learning will freeze the language model and directly perform few-shot or zero-shot learning on it. 
 This enables the language models to be pre-trained on large amount of raw text data and be adapted to new domains without tuning it again. Hence, prompt learning could help save much time and efforts.
 \subsubsection{Traditional Prompt Learning}
 The process of utilizing prompt learning with a language model can be divided into two main stages: prompt engineering and answer engineering.
 
\begin{itemize}
    \item {Prompt engineering.} In general, there are two commonly used forms of prompt engineering: discrete prompt and continuous prompt. Discrete prompts are typically manually designed by humans for specific tasks, while continuous prompts are added to the input embeddings to convey task-specific information.
    \item {Answer engineering.} After the task has been reformulated, the answer generated by the language model based on the provided prompt needs to be mapped to the ground truth space. There are different paradigms for answering engineering, including discrete search space and continuous search space. As this topic is more closely related to classification tasks, we refer interested readers to  for further information.
\end{itemize}

In addition to single-prompt learning methods, there are also multi-prompt methods. These approaches primarily focus on ensembling multiple prompts together as input during inference to improve prediction robustness, which is more effective than relying on a single prompt. Another approach to multi-prompt learning is prompt augmentation, which aims to assist the model in answering questions by providing additional prompts that have already been answered.

\subsubsection{In-context Learning}
Recently, in-context learning has received significant attention as an effective method for improving language models' performance. This approach is a subset of prompt learning and involves using a pre-trained language model as the backbone, along with adding a few input-label demonstration pairs and instructions to the prompt. In-context learning has been shown to be highly effective in guiding language models to produce better answers that are more closely aligned with the given prompt. Some recent studies have also suggested that in-context learning can be viewed as a form of implicit fine-tuning, as it enables the model to learn how to generate answers more accurately based on the input prompt.


\subsection{Efficiency in Pretrained Foundation Models}
Within the context of the AIGC framework, a fundamental component of each proposed method involves utilizing large pretrained foundation models (PFMs)~\cite{zhou2023comprehensive}. PFMs, such as BERT~\cite{devlin2018bert}, GPT-2~\cite{Radford2019LanguageMA}, and RoBERTa~\cite{liu2019roberta}, have revolutionized the field of natural language processing by achieving state-of-the-art results on a wide range of NLP tasks. However, these models are incredibly large and computationally expensive, which can lead to efficiency problems. This is especially true when working with limited computational resources, such as on personal computers or in cloud environments with limited processing power. In order to address these efficiency problems,
recent numerous works have been dedicated to exploring more cost-effective pretraining methods to pretrain large-scale PFMs. The effectiveness of learning algorithms is contingent upon both training methods and model architecture efficiency. For example, ELECTRA~\cite{clark2020electra} introduces an RTD task that predicts whether each input marker is replaced by other tokens, thereby enabling ELECTRA to train against all input tokens. In addition to effective training methods, model architecture efficiency can also contribute to improved PFMs efficiency. Most PFMs based on the Transformer algorithm may benefit from a more efficient model architecture by reducing the complexity of the Transformer algorithm.

\subsection{Model Compression}
Model compression is an effective approach to reduce model size and improve computation efficiency. It requires fewer computing resources and memory and can better meet the needs of various applications than the original model, where its strategies can be divided into two categories: parameter compression and structure compression. Parameter compression methods include parameter pruning, parameter quantization, low-rank decomposition, and parameter sharing. Parameter pruning deletes redundant parameters based on a sizeable PFM, while parameter quantization reduces model parameters to lower-order numbers without significant impact on model performance. Low-rank decomposition reduces the dimension of a high-dimensional parameter vector, and parameter sharing maps model parameters to reduce their number. Structure compression refers to designing new compact network structures and employing knowledge distillation, where the knowledge learned from a larger teacher model is transferred to a smaller student model through soft labels, among other techniques. DistilBERT~\cite{sanh2019distilbert}, for instance, uses knowledge distillation to compress BERT, reducing its size by 40\% while maintaining 97\% of its language comprehension. ALBERT uses decomposition-embedded parameterization and cross-layer parameter sharing to reduce the number of model parameters.

\section{Trustworthy \& Responsible AIGC}\label{sec:7}
While AIGC has the potential to be incredibly useful in many different applications, it also raises significant concerns about security and privacy. In this section, we will discuss studies that disclose the "dark" side of AIGC and countermeasures proposed to ensure that AIGC can be used in a safe and responsible way.

\subsection{Security}

\header{Factuality} 
Although tools like ChatGPT \cite{ChatGPT_2022} is capable of generating content that usually appears or sounds reasonable, they are often unreliable in terms of factuality \cite{chatgpt_failures}. Sometimes, the model outputs counterfactual or even absurd answers, which pose a serious threat to the truthfulness of information on the internet. Recently, NewsGuard's Misinformation Monitor \cite{newsguard_2023} has indicated the possibility that AI-generated content tools are being weaponized to spread misinformation at an unprecedented scale. Presented with 100 samples from NewsGuard's proprietary misinformation database, the tested model, ChatGPT, generated false narratives for 80 of the 100 previously identified false arguments, which could easily come across as legitimate and authoritative for those unfamiliar with the topics \cite{newsguard_2023}. Moreover, Alex \cite{mahadevan_2023} offers a more specific example by demonstrating how to leverage ChatGPT \cite{ChatGPT_2022} to generate a newspaper. Besides natural language processing, factuality concerns also exist in the computer vision domain. For example, stable diffusion \cite{rombach_high-resolution_2022}, which has been demonstrated to be a powerful vision-generated model, has trouble drawing realistic human hands with the correct number of fingers \cite{dataconomy_2023}. To prevent the spread of misinformation on the internet, websites like Stackoverflow \cite{wang2013empirical} propose policies that ban users from using AI-generated content as an answer to reduce the risk of being overwhelmed by inaccurate and biased content.

Earlier studies have shown that AI models suffer from factual incorrectness and hallucination of knowledge \cite{zhang2018texttruth}. To evaluate and improve the factual accuracy of AI-generated content, \cite{goodrich2019assessing} proposed model-based metrics that measure the factualness of generated text, complementing traditional metrics such as ROUGE (Recall-Oriented Understudy for Gisting Evaluation) \cite{lin2004rouge} and BLEU (Bilingual Evaluation Understudy) \cite{papineni2002bleu}. Specifically, \cite{goodrich2019assessing} proposed a Transformer-based end-to-end fact extraction model, which enables the structured prediction of relation tuples for factualness assessment. 
More systematic definitions of truthfulness standards and approaches for governing AI-generated content were later proposed in Truthful AI \cite{evans2021truthful}. The standard proposed by Truthful AI aims to avoid "negligent falsehoods" and explicitly train AI systems to be truthful via curated datasets and human interaction. 
Based on GPT-3, WebGPT \cite{nakano2021webgpt} proposed a humanoid prototype that models the AI answering process into web searching and evidence-composing phrases. Since the model is trained to cite its sources, the factual accuracy of AI-generated content is significantly improved in multiple benchmark datasets \cite{fan2019eli5, lin2021truthfulqa}. Specifically, the model is obtained by fine-tuning GPT-3 using imitation learning, which leverages human feedback to optimize answer quality. 
Furthermore, \cite{lee2022factuality} measures and improves the factual accuracy of large-scale language models for open-ended text generation. \cite{lee2022factuality} proposed the \textit{factual-nucleus} sampling algorithm that dynamically adapts the randomness to balance the factuality and quality of AI-generated content. A \textit{factuality-enhanced} training method that uses \textit{TOPICPREFIX} for better awareness of facts and sentence completion is designed as the training objective, which vastly reduces factual errors.
Despite these preliminary advances in developing more truthful AI, challenges still remain. For example, AI-generated content might be problematic on unfamiliar types of questions and contexts that involve contradictions \cite{chatgpt_failures}.

\header{Toxicity}
Besides utility, it is important for AI-generated content (AIGC) to be helpful, harmless, unbiased, and non-toxic. Extensive research has been conducted on the potential harm caused by deployed models \cite{bender2021dangers, bommasani2021opportunities, kenton2021alignment}, which can include biased outputs \cite{dhamala2021bold, liang2021towards}, stereotypes \cite{nadeem2020stereoset}, and misinformation \cite{solaiman2019release}. To address this issue of toxicity in the language domain, OpenAI proposes InstructGPT \cite{ouyang_training_2022}, which aligns language models with human preferences by using human feedback as a reward signal to fine-tune the models, ensuring more relevant and safe responses.
Concurrently, Google proposes LaMDA \cite{thoppilan2022lamda}, a family of neural language models specialized for safe and factual dialog by leveraging fine-tuning and external knowledge sources. To improve model safety, LaMDA \cite{thoppilan2022lamda} designs a set of metrics (Appendix A.1 in the original paper) that quantify model safety based on an illustrative set of human values derived from Google's AI Principles \footnote{https://ai.google/principles/}. 
Furthermore, Ganguli \etal \cite{ganguli2022red} study and improve the safety of language models in an adversarial way. Specifically, they investigate the scaling behaviors for red teaming across models with different sizes (2.7B, 13B, and 52B parameters) and training schemes (plain LM, fine-tuned LM, LM with rejection sampling, and LM trained with RLHF). They found that models trained with RLHF scale better and are increasingly difficult to red team.



\subsection{Privacy}
\header{Membership inference} The goal of the membership inference attack (MIA) is to determine whether an image $x$ belongs to the set of training data. Wu \textit{et al.} \cite{wu2022membership} investigated the membership leakage in text-to-image (diffusion-based and sequence-to-sequence-based) generation models under realistic black-box settings. Specifically, three kinds of intuitions including quality, reconstruction error, and faithfulness are considered to design the attack algorithms. However, Wu \textit{et al.} \cite{wu2022membership} assumed that the member set and the hold-out set come from different distributions, which makes the MIA much easier. Under a more practical setting \cite{hu2022membership}, where the member set and the hold-out set are in the same distribution, Duan \textit{et al.} \cite{duan2023diffusion} propose Step-wise Error Comparing Membership Inference (SecMI), a black-box MIA that infers memberships by assessing the matching of forward process posterior estimation at each timestep. Concurrently, Hu and Pang \cite{hu2023membership} propose two attack approaches, including loss-based and likelihood-based MIA. Furthermore, Matsumoto \textit{et al.} \cite{matsumoto2023membership} introduce more comparisons with GANs.
\header{Data Extraction} 
The objective of a data extraction attack is to retrieve an image from the set of training data, denoted as $x \in D$. The attack can be considered a success if the attacker is able to obtain an image $\hat{x}$ that closely resembles image $x \in D$. Compared to the membership inference attack, the data extraction attack poses stronger privacy risks to the model. The feasibility of such an attack might be due to the memorization property of large-scale models \cite{carlini2022quantifying}, in which they turn to memorize parts of their training data. When prompted appropriately, the memorized training data that might contain sensitive information will be emitted verbatim. 
Earlier, in the language domain, Carlini \etal \cite{carlini2021extracting} demonstrated that large language models (specifically, GPT-2 \cite{radford2019language}) memorize and leak individual training examples. Specifically, they proposed a simple and efficient method for extracting verbatim sequences from a language model’s training set using only black-box query access.
Recently, in the vision domain, Somepalli \etal \cite{somepalli2021diffusion} showed that the data replication problem existed in diffusion models, where the generated images are close to the training data in terms of semantic similarity. To disclose worse-case privacy risk, Carlini \etal \cite{carlini2023extracting} further explored the privacy vulnerabilities of state-of-the-art diffusion models by leveraging a generate-and-filter pipeline to extract over a thousand training examples from the models. Specifically, the extraction approach first samples 500 candidate images by querying the generation function in a black-box manner using selected prompts. Based on the intuition that generations of memorized data are nearly identical, a similarity graph is then constructed to determine whether an image belongs to the training set. The results in \cite{carlini2023extracting} show that diffusion models, including Stable Diffusion \cite{rombach_high-resolution_2022} and Imagen \cite{saharia2022photorealistic}, are more susceptible to privacy breaches compared to earlier generative models like GANs \cite{goodfellow2014gan}. These results highlight the necessity of developing new techniques for preserving privacy during training to address these vulnerabilities. 

\section{Open Problems and Future Directions}\label{sec:8}

In this section, we discuss some challenges in AIGC and potential ways to address them.

\textbf{High-stakes Applications.} Although the community has witnessed the huge success of AIGC in images, texts, and audio generations, these areas are arguably more fault-tolerant. On the contrary, AIGC for high-stakes applications, including healthcare~\cite{reddy2020governance}, financial services~\cite{qi2018fintech}, autonomous vehicles~\cite{grigorescu2020survey}, and science discovery~\cite{gil2014amplify},  are still challenging. In these domains, tasks are mission-critical and require a high degree of accuracy, reliability, transparency, and less or near zero fault-tolerant. For example, the large language model Galactica~\cite{taylor2022galactica}, which is made for automatically organizing science, can perform knowledge-intensive scientific tasks and have promising performances on several benchmark tasks. Its public demo were taken down from the service only three days after its initial release, due to the intensive criticism on its generated biased and incorrect results in an authoritative tone. It would be crucial for generative models in these high-stakes applications to give confidence scores, reasoning, and source information along with generated results. Only when professionals understand how and where these results are coming from, they can confidently utilize these tools in their tasks.

\textbf{Specialization and Generalization.}  AIGC relies on the choice of foundation models, which are trained on different datasets, including crawl-based~\cite{clip2021} one and carefully curated~\cite{taylor2022galactica}. And it is argued in~\cite{bommasani2021opportunities} that ``training on a more diverse dataset is not always better for downstream performance than a more specialized foundation model."  However, the curation of highly specialized dataset can be both time-consuming and cost-ineffective.
A better understanding of cross-domain representations and how they are resilient to testing-time distribution-shift may guide the design of training datasets that balance specialization and generalization.

\textbf{Continual Learning and Retraining.} The human knowledge base keeps expanding and new tasks continue emerging. To generate the contents with up-to-date information, it not only requires model to ``remember" the learned knowledge, but also be able to learn and infer from newly acquired information. For some scenarios~\cite{ostapenko2022continual}, it suffices to perform the continual learning on downstream tasks while keeping the pre-trained foundation model unchanged. When necessary~\cite{gururangan2020don}, one can perform continual learning on foundation models. However, it is also observed that the continual learning may not always outperform the retrained models~\cite{chalkidis2020legal}.
This calls for the need to understand when should one choose continual learning strategy and when to choose to the retraining strategy. Also, training foundation models from scratch may be prohibitive, so modularized design of next generation of foundation models for AIGC may elucidate which parts of the model should be retrained.

\textbf{Reasoning.}
Reasoning is a crucial component of human intelligence that enables us to draw inferences, make decisions, and solve complex problems. However, even trained with large scale dataset, sometimes GAI models could still fail at common sense reasoning tasks~\cite{wei2022chain,zhang2023multimodal}. 
Recently, more and more researchers began to focus on this problem. 
Chain-of-thought (CoT) prompting~\cite{wei2022chain} is a promising solution to the challenge of reasoning in generative AI models. It is designed to enhance the ability of large language models to learn about logical reasoning in the context of question answering.
By explaining the logical reasoning process that humans use to arrive at answers to models, they can follow the same road that humans take in processing their reasoning.
 By incorporating this approach, large language models can achieve higher accuracy and better performance in tasks that require logical reasoning.
CoT has also been applied to other areas like vision language question answering~\cite{zhang2023multimodal} and code generation~\cite{madaan2022language}.
However, it still remains a problem that how to construct these CoT prompts according to specific tasks.

\textbf{Scaling up.}
Scaling up has been a common problem in large-scale pretraining. Model training is always limited by compute budget, available dataset and model size. As the size of pretraining models increases, the time and resources required for training also increases significantly. This poses a challenge for researchers and organizations that seek to utilize large-scale pretraining for various tasks, such as natural language understanding, computer vision, and speech recognition.
Another issue pertains to the efficacy of pretraining with large-scale datasets, which may not yield optimal results if experimental hyperparameters, such as model size and data volume, are not thoughtfully designed. As such, suboptimal hyperparameters can result in wasteful resource consumption and the failure to achieve desired outcomes through further training.
Several works have been proposed to solve these problems. 
Hoffmann et al.~\cite{hoffmann2022training} introduce a formal scaling law to predict model performance based on the number of parameters and dataset size. This work provides a useful framework for understanding the relationship between these key factors when scaling up. 
Aghajanyan et al.~\cite{aghajanyan2023scaling} conduct empirical analyses to validate the Hoffmann scaling law and propose an additional formula that explores the relationship between different training tasks in multimodal model training settings. These findings provide valuable insights into the complexities of large-scale model training and the nuances of optimizing performance across diverse training domains.

\textbf{Social issues.}
As AIGC continues to proliferate across various domains, social concerns regarding its use have become increasingly prominent. These concerns relate to issues such as bias, ethics, and the impact of AI-generated content on various stakeholders. One major concern is the potential for bias in AI-generated content, particularly in areas such as natural language processing and computer vision. AI models can inadvertently perpetuate or amplify existing societal biases, particularly if the training data used to develop the models are themselves biased. This can have significant negative consequences, such as perpetuating discrimination and inequities in areas such as hiring, loan approvals, and criminal justice. Ethical concerns also arise with the use of AI-generated content, particularly in cases where the technology is used to generate deepfakes or other forms of manipulated media. Such content can be used to spread false information, incite violence, or harm individuals or organizations. Additionally, there are concerns around the potential for AI-generated content to infringe on copyright and intellectual property rights, as well as issues around privacy and data security. Overall, while AI-generated content holds significant promise in various domains, it is crucial to address these social concerns to ensure that its use is responsible and beneficial for society as a whole.

\section{Conclusion} \label{sec:9}
This survey provides a comprehensive overview of the history and recent advancements in AIGC, with a particular focus on both unimodality and multimodality generative models. In addition, we discuss the recent applications of generative AI models, commonly used techniques in AIGC, and address concerns surrounding trustworthiness and responsibility in the field. Finally, we explore open problems and future directions for AIGC, highlighting potential avenues for innovation and progress.
The primary objective of this survey is to provide readers with a comprehensive understanding of recent developments and future challenges in generative AI. Our analysis of the general framework of AI generation aims to distinguish contemporary generative AI models from their predecessors. Ultimately, we hope this survey will aid readers in gaining deeper insights into this field.
Moving forward, we would further investigate this topic and provide a more comprehensive analysis of AIGC.


\bibliographystyle{ieeetr}
\bibliography{sample-base}

\appendix
\section{Curated Advances in Generative AI}
In this section, we present a comprehensive review of the recent significant advancements in the field of generative AI. 
Consistent with the aforementioned discussion, we classify the general framework into unimodal and multimodal generative models.
In each subsection, we further categorize the models based on specific modalities, and we provide a table summary of relative paper details.
\subsection{Language}

In this section, we give a summary of the main milestone models in NLP. Generally, the architecture includes probabilistic objectives, encoder, decoder, and encoder-decoder structure. We also summarize the backbones of these methods.

\begin{table}[H]
    \centering
    \resizebox{\textwidth}{!}{
    \begin{tabular}{llllll}
    \hline
    \textbf{Year} & \textbf{Conference} & \textbf{Method} & \textbf{Architecture} & \textbf{Backbone} & \textbf{Code}   
     \\ \hline    
2013 & NeurIPS & Skip-Gram~\cite{mikolov2013distributed}      & Probabilistic   & Word2Vec    & \href{https://github.com/tensorflow/models}{https://github.com/.../models}                                                                        \\
2014 & EMNLP   & GloVe~\cite{pennington2014glove}          & Probabilistic   & Word2Vec    & \href{https://github.com/stanfordnlp/GloVe}{https://github.com/.../GloVe}                                                                         \\
2015 & NeurIPS & LM-LSTM~\cite{dai2015semi}        & Probabilistic   & LSTM        & -                                                                                                                                                 \\
2017 & TACL    & FastText~\cite{bojanowski2017enriching}       & Probabilistic   & Word2Vec    & \href{https://github.com/facebookresearch/fastText}{https://github.com/.../fastText}                                                              \\
2018 & NAACL   & ELMO~\cite{peters2018deep}           & Encoder         & LSTM        & \href{https://allennlp.org/elmo}{https://allennlp.org/elmo}                                                                                       \\
2018 & -       & GPT~\cite{Radford2018ImprovingLU}            & Decoder         & Transformer & \href{https://github.com/huggingface/transformers}{https://github.com/huggingface/transformers}                                                   \\
2019 & ACL     & ERNIE~\cite{sun2019ernie}          & Encoder         & Transformer & \href{https://github.com/PaddlePaddle/ERNIE}{https://github.com/.../ERNIE}                                                                        \\
2019 & ACL     & Transformer-XL~\cite{dai2019transformer} & Encoder         & Transformer & \href{https://github.com/kimiyoung/transformer-xl}{https://github.com/.../transformer-xl}                                                         \\
2019 & NeurIPS & UNILM~\cite{dong2019unified}          & Encoder         & Transformer & \href{https://github.com/microsoft/unilm}{https://github.com/.../unilm}                                                                           \\
2019 & NAACL   & BERT~\cite{devlin2018bert}           & Encoder         & Transformer & \href{https://github.com/google-research/bert}{https://github.com/google-research/bert}                                                           \\
2019 & CoRR    & RoBERTa~\cite{liu2019roberta}        & Encoder         & Transformer & \href{https://github.com/pytorch/fairseq}{https://github.com/pytorch/fairseq}                                                                     \\
2019 & NeurIPS & XLNet~\cite{yang2019xlnet}          & Encoder         & Transformer & \href{https://github.com/zihangdai/xlnet}{https://github.com/zihangdai/xlnet}                                                                     \\
2019 & NeurIPS & DistilBERT~\cite{sanh2019distilbert}     & Encoder         & Transformer & \href{https://github.com/huggingface/transformers}{https://github.com/huggingface/transformers}                                                   \\
2019 & ICML    & MASS~\cite{song2019mass}           & Encoder         & Transformer & \href{https://github.com/microsoft/MASS}{https://github.com/microsoft/MASS}                                                                       \\
2019 & ICLR    & StructBERT~\cite{wang2019structbert}        & Encoder         & Transformer & -    \\
2019 & EMNLP    & KnowBERT~\cite{peters2019knowledge}        & Encoder         & Transformer & \href{https://github.com/allenai/kb}{https://github.com/.../kb}     \\
2019 & -       & GPT-2~\cite{radford2019language}          & Decoder         & Transformer & \href{https://github.com/openai/gpt-2}{https://github.com/openai/gpt-2}                                                                           \\
2019 & JMLR    & T5~\cite{raffel2020exploring}             & Encoder-Decoder & Transformer & \href{https://github.com/google-research/text-to-text-transfer-transformer}{https://github.com/google-research/text-to-text-transfer-transformer} \\
2019 & -       & Megatron~\cite{megatron}       & General         & Transformer & \href{https://github.com/NVIDIA/Megatron-LM}{https://github.com/NVIDIA/Megatron-LM}                                                               \\
2020 & ACL     & fastBERT~\cite{liu2020fastbert}       & Encoder         & Transformer & \href{https://github.com/autoliuweijie/FastBERT}{https://github.com/.../FastBERT}                                                                 \\
2020 & ACL     & spanBERT~\cite{joshi2020spanbert}       & Encoder         & Transformer & \href{https://github.com/facebookresearch/SpanBERT}{https://github.com/.../SpanBERT}                                                              \\
2020 & ICLR    & Reformer~\cite{kitaev2020reformer}       & Encoder         & Reformer    & \href{https://github.com/google/trax/tree/master/trax/models/reformer}{https://github.com/.../reformer}                                           \\
2020 & EMNLP   & TinyBERT~\cite{jiao2019tinybert}       & Encoder         & Transformer & \href{https://github.com/huawei-noah/Pretrained-Language-Model/tree/master/TinyBERT}{https://github.com/.../TinyBERT}                              \\
2020 & ICLR    & ALBERT~\cite{lan2019albert}         & Encoder         & Transformer & \href{https://github.com/google-research/ALBERT}{https://github.com/google-research/ALBERT}                                                       \\
2020 & ICLR    & ELECTRA~\cite{clark2020electra}        & Encoder         & Transformer & \href{https://github.com/google-research/electra}{https://github.com/google-research/electra}                                                     \\
2020 & NeurIPS & GPT-3~\cite{radford2019language}          & Decoder         & Transformer & \href{https://github.com/openai/gpt-3}{https://github.com/openai/gpt-3}                                                                           \\
2020 & ACL     & BART~\cite{lewis2019bart}           & Encoder-Decoder & Transformer & \href{https://github.com/huggingface/transformers}{https://github.com/huggingface/transformers}                                                   \\
2020 & -       & PaLM~\cite{chowdhery2022palm}           & Decoder         & Transformer & \href{https://github.com/lucidrains/PaLM-pytorch}{https://github.com/lucidrains/PaLM-pytorch}                                                     \\
2021 & -       & Gopher~\cite{rae2021scaling}         & Decoder         & Transformer & -                                                                                                                                                 \\
2021 & JMLR    & Switch~\cite{fedus2021switch}         & Encoder-Decoder & Transformer & \href{https://github.com/tensorflow/mesh}{https://github.com/tensorflow/mesh}                                                                     \\
2022 & -       & LaMDA~\cite{thoppilan2022lamda}          & Decoder         & Transformer & \href{https://github.com/conceptofmind/LaMDA-rlhf-pytorch}{https://github.com/.../LaMDA}                                                          \\
2022 & -       & OPT~\cite{opt}            & Decoder         & Transformer & \href{https://github.com/facebookresearch/metaseq}{https://github.com/facebookresearch/metaseq}                                                   \\
2022 & -       & InstructGPT~\cite{ouyang_training_2022}    & Decoder         & Transformer & \href{https://github.com/openai/following-instructions-human-feedback}{https://github.com/openai/following-instructions-human-feedback}           \\
2022 & -       & Sparrow~\cite{glaese2022improving}        & Decoder         & Transformer & -                                                                                                                                                 \\
2022 & -       & BLOOM~\cite{scao2022bloom}          & Decoder         & Transformer & \href{https://github.com/bigscience-workshop/bigscience}{https://github.com/bigscience-workshop/bigscience}                                       \\
2022 & -       & MT-NLG~\cite{mt-nlg}         & Decoder         & Transformer & \href{https://github.com/microsoft/DeepSpeed}{https://github.com/microsoft/DeepSpeed}                                                             \\
2022 & ICLR    & HTLM~\cite{html}           & Encoder-Decoder & Transformer & -                                                                                                                                                 \\
2022 & ACL     & DQ-BART~\cite{li2022dq}        & Encoder-Decoder & Transformer & \href{https://github.com/amazon-research/dq-bart}{https://github.com/amazon-research/dq-bart}                                                     \\
2022 & ICLR    & ExT5~\cite{aribandi2021ext5}           & Encoder-Decoder & Transformer & \href{https://github.com/google-research/text-to-text-transfer-transformer}{https://github.com/google-research/text-to-text-transfer-transformer} \\
2023 & -       & LLaMA~\cite{touvron2023llama}          & Decoder         & Transformer & -                                                                                                                                                                                                                                              

\\ \hline
    \end{tabular}}
    \caption{Major natural language models.}
    \label{tab:unimodal_nlg}
\end{table}

\subsection{Vision}
\begin{table}[H]
    \centering
    \resizebox{0.9\textwidth}{!}{
    \begin{tabular}{lllll}
    \hline
\textbf{Year} & \textbf{Method}                 & \textbf{Architecture} & \textbf{Category}& \textbf{Code}     
     \\ \hline    
2014 & GAN~\cite{goodfellow2014gan}                     & GAN       & Traditional                 & \href{https://github.com/goodfeli/adversarial}{https://github.com/goodfeli/adversarial}                                                              \\
2015 & LAPGAN~\cite{denton2015deep}                     & GAN       & Traditioal                  & \href{https://github.com/facebook/eyescream}{https://github.com/facebook/eyescream}                                                                  \\
2015 & DCGANs~\cite{radford2015unsupervised}            & GAN       & Traditioal                  & \href{https://github.com/carpedm20/DCGAN-tensorflow}{https://github.com/carpedm20/DCGAN}                                                  \\
2017 & Progressive GAN ~\cite{karras2017progressive}    & GAN       & Traditioal                  & \href{https://github.com/tkarras/progressive_growing_of_gans}{https://github.com/tkarras/progre...}                                                  \\
2019 & SAGAN ~\cite{zhang2019self}                      & GAN       & Traditioal                  & \href{https://github.com/brain-research/self-attention-gan}{https://github.com/brain-research/self}                                    \\
2018 & BigGAN~\cite{brock2018large}                     & GAN       & Traditioal                  & \href{https://github.com/ajbrock/BigGAN-PyTorch}{https://github.com/ajbrock/BigGAN}                                                               \\
2019 & StyleGAN~\cite{karras2019style}                  & GAN       & Traditioal                  & \href{https://github.com/NVlabs/stylegan}{https://github.com/NVlabs/stylegan}                                                                        \\
2016 & BiGAN~\cite{donahue2016adversarial}              & GAN       & Traditioal                  & \href{https://github.com/eriklindernoren/Keras-GAN}{https://github.com/eriklindernoren/GAN}                                                    \\
2018 & AGE~\cite{ulyanov2018takes}                      & GAN       & Traditioal                  & \href{https://github.com/DmitryUlyanov/AGE}{https://github.com/DmitryUlyanov/AGE}                                                                    \\
2017 & D2GAN~\cite{nguyen2017dual}                      & GAN       & Traditioal                  & \href{https://github.com/tund/D2GAN}{https://github.com/tund/D2GAN}                                                                                  \\
2016 & GMAN~\cite{durugkar2016generative}               & GAN       & Traditioal                  & \href{https://github.com/zhengchuanpan/GMAN}{https://github.com/zhengchuanpan/GMAN}                                                                  \\
2017 & MGAN~\cite{hoang2017multi}                       & GAN       & Traditioal                  & \href{https://github.com/qhoangdl/MGAN}{https://github.com/qhoangdl/MGAN}                                                                            \\
2018 & MAD-GAN~\cite{ghosh2018multi}                    & GAN       & Traditioal                  & \href{https://github.com/LiDan456/MAD-GANs}{https://github.com/LiDan456/MAD-GANs}                                                                    \\
2016 & CoGAN~\cite{liu2016coupled}                      & GAN       & Traditioal                  & \href{https://github.com/mingyuliutw/CoGAN}{https://github.com/mingyuliutw/CoGAN}                                                                    \\
2016 & InfoGAN~\cite{chen2016infogan}                   & GAN       & Representative variants     & \href{https://github.com/eriklindernoren/PyTorch-GAN}{https://github.com/eriklindernoren/...GAN}                                                     \\
2014 & CGANs~\cite{mirza2014conditional}                & GAN       & Representative variants     & \href{https://github.com/pfnet-research/sngan_projection}{https://github.com/pfnet-research/sngan...}                                                \\
2018 & C-CycleGAN~\cite{lu2018attribute}                & GAN       & Representative variants     & -                                                                                                                                                    \\
2019 & MSGAN~\cite{mao2019mode}                         & GAN       & Representative variants     & \href{https://github.com/HelenMao/MSGAN}{https://github.com/HelenMao/MSGAN}                                                                          \\
2016 & f-GAN~\cite{nowozin2016f}                        & GAN       & Representative variants     & \href{https://github.com/mboudiaf/Mutual-Information-Variational-Bounds}{https://github.com/mboudiaf/Mut...}                                         \\
2017 & WGAN~\cite{gulrajani2017improved}                & GAN       & Objective                   & \href{https://github.com/daheyinyin/wgan}{https://github.com/daheyinyin/wgan}                                                                        \\
2020 & GLS-GAN~\cite{qi2020loss}                        & GAN       & Objective                   & \href{https://github.com/guojunq/lsgan}{https://github.com/guojunq/lsgan}                                                                            \\
2017 & LS-GAN~\cite{mao2017least}                       & GAN       & Objective                   & \href{https://github.com/xudonmao/LSGAN}{https://github.com/xudonmao/LSGAN}                                                                          \\
2018 & SNGAN~\cite{miyato2018spectral}                  & GAN       & Objective                   & \href{https://github.com/pfnet-research/sngan_projection}{https://github.com/pfnet-research/sngan}                                                   \\
2016 & Che et al.~\cite{che2016mode}                    & GAN       & Objective                   & -                                                                                                                                                    \\
2016 & UnrolledGAN~\cite{metz2016unrolled}              & GAN       & Objective                   & \href{https://github.com/poolio/unrolled_gan}{https://github.com/poolio/unrolledgan}                                                                \\
2018 & RelativisticGAN~\cite{jolicoeur2018relativistic} & GAN       & Objective                   & \href{https://github.com/AlexiaJM/RelativisticGAN}{https://github.com/AlexiaJM/RelativisticGAN}                                                      \\
2013 & VAE~\cite{kingma2013auto}                        & VAE       & Traditional                 & \href{https://github.com/AntixK/PyTorch-VAE}{https://github.com/AntixK/PyTorch-VAE}                                                                  \\
2018 & VampPrior~\cite{tomczak2018vae}                  & VAE       & Complex priors              & \href{https://github.com/jmtomczak/vae_vampprior}{https://github.com/jmtomczak/vaevampprior}                                                        \\
2019 & BIVA\cite{maaloe2019biva}                        & VAE       & Complex priors              & \href{https://github.com/vlievin/biva-pytorch}{https://github.com/vlievin/biva-pytorch}                                                              \\
2020 & NVAE\cite{vahdat2020nvae}                        & VAE       & Complex priors              & \href{https://github.com/NVlabs/NVAE}{https://github.com/NVlabs/NVAE}                                                                                \\
2021 & GHVAE~\cite{wu2021greedy}                        & VAE       & Complex priors              & -                                                                                                                                                    \\
2019 & RAE~\cite{ghosh2019variational}                  & VAE       & Regularized Autoencoders    & \href{https://github.com/ParthaEth/Regularized_autoencoders-RAE-}{https://github.com/ParthaEth/Regul...}                                             \\
2021 & dGAN~\cite{rolfe2016discrete}                    & VAE       & Regularized Autoencoders    & \href{https://github.com/topics/discrete-variational-autoencoders}{https://github.com/topics/discrete...}                                            \\
2017 & VQ-VAE~\cite{van2017neural}                      & VAE       & Regularized Autoencoders    & \href{https://github.com/deepmind/sonnet/blob/v2/sonnet/src/nets/vqvae.py}{https://github.com/deepmind/...vqvae}                                     \\
2019 & VQ-VAE2~\cite{razavi2019generating}              & VAE       & Regularized Autoencoders    & \href{https://github.com/deepmind/sonnet}{https://github.com/deepmind/sonnet}                                                                        \\
2014 & NICE~\cite{dinh2014nice}                         & Flow      & Coupling and autoregressive & \href{https://github.com/EugenHotaj/pytorch-generative/blob/master/pytorch_generative/models/flow/nice.py}{https://github.com/EugenHotaj/...nice} \\
2016 & Real NVP~\cite{dinh2016density}                  & Flow      & Coupling and autoregressive & \href{https://github.com/tensorflow/models}{https://github.com/tensorflow/models}                                                                    \\
2017 & MAF~\cite{papamakarios2017masked}                & Flow      & Coupling and autoregressive & \href{https://github.com/gpapamak/maf}{https://github.com/gpapamak/maf}                                                                              \\
2018 & NAF~\cite{huang2018neural}                       & Flow      & Coupling and autoregressive & \href{https://github.com/CW-Huang/NAF}{https://github.com/CW-Huang/NAF}                                                                              \\
2020 & BNAF~\cite{de2020block}                          & Flow      & Coupling and autoregressive & \href{https://github.com/nicola-decao/BNAF}{https://github.com/nicola-decao/BNAF}                                                                    \\
2017 & ConvFlow~\cite{zheng2017convolutional}           & Flow      & Convolutional and Residual  & -                                                                                                                                                    \\
2019 & E-ConvFlow~\cite{hoogeboom2019emerging}          & Flow      & Convolutional and Residual  & \href{https://github.com/ehoogeboom/emerging}{https://github.com/ehoogeboom/emerging}                                                                \\
2017 & RevNets~\cite{gomez2017reversible}               & Flow      & Convolutional and Residual  & \href{https://github.com/renmengye/revnet-public}{https://github.com/renmengye/revnet-public}                                                        \\
2018 & i-RevNets~\cite{jacobsen2018revnet}              & Flow      & Convolutional and Residual  & \href{https://github.com/jhjacobsen/pytorch-i-revnet}{https://github.com/jhjacobsen/pytorch-i-revnet}                                                \\
2020 & DDPM~\cite{ho2020denoising}                      & Diffusion & Traditional                 & \href{https://github.com/hojonathanho/diffusion}{https://github.com/hojonathanho/diffusion}                                                          \\
2019 & NCSN~\cite{song2019DSM}                          & Diffusion & Traditional                 & \href{https://github.com/ermongroup/ncsn}{https://github.com/ermongroup/ncsn}                                                                        \\
2020 & Score SDE~\cite{song2020score}                   & Diffusion & Training Enhance            & \href{https://github.com/yang-song/score_sde}{https://github.com/yang-song/scoresde}                                                                \\
2022 & Salimans et al.~\cite{salimans2022progressive}   & Diffusion & Training Enhance            & \href{https://github.com/Hramchenko/diffusion_distiller}{https://github.com/Hramchenko/diffusion..distiller}                                          \\
2022 & TDPM~\cite{zheng2022truncated}                   & Diffusion & Training Enhance            & \href{https://github.com/jegzheng/truncated-diffusion-probabilistic-models}{https://github.com/jegzheng/truncat..}                                   \\
2022 & ES-DDPM~\cite{lyu2022accelerating}               & Diffusion & Training Enhance            & \href{https://github.com/zhaoyanglyu/early_stopped_ddpm}{https://github.com/zhaoyanglyu/early...}                                                    \\
2022 & Franzese et al.~\cite{2206.05173}                & Diffusion & Training Enhance            & -                                                                                                                                                    \\
2021 & Improved DDPM~\cite{nichol2021improved}          & Diffusion & Training Enhance            & \href{https://github.com/openai/improved-diffusion}{https://github.com/openai/improved-diffusion}                                                    \\
2021 & San Roman et al.~\cite{san2021noise}             & Diffusion & Training Enhance            & -                                                                                                                                                    \\
2020 & DDIM~\cite{song2020denoising}                    & Diffusion & Training-free Sampling      & \href{https://github.com/ermongroup/ddim}{https://github.com/ermongroup/ddim}                                                                        \\
2022 & Analytic-DPM~\cite{bao2022analytic}              & Diffusion & Training-free Sampling      & \href{https://github.com/baofff/Analytic-DPM}{https://github.com/baofff/Analytic-DPM}                                                                \\
2021 & Watson et al.~\cite{watson2021learning}          & Diffusion & Training-free Sampling      & -                                                                                                                                                    \\
2022 & Watson et al.~\cite{watson2022learning}          & Diffusion & Training-free Sampling      & -                                                                                                                                                    \\
2021 & Nachmani et al.~\cite{2106.07582}                & Diffusion & Noise Distribution          & -                                                                                                                                                    \\
2022 & Cold Diffusion~\cite{bansal2022cold}             & Diffusion & Noise Distribution          & \href{https://github.com/arpitbansal297/cold-diffusion-models}{https://github.com/arpitbansal297/cold-...}                                           \\
2021 & CCDF~\cite{2112.05146}                           & Diffusion & Noise Distribution          & -                                                                                                                                                    \\
2022 & DiffuseVAE~\cite{2201.00308}                     & Diffusion & Mixed Modeling              & \href{https://github.com/kpandey008/DiffuseVAE}{https://github.com/kpandey008/DiffuseVAE}                                                            \\
2021 & LSGM~\cite{vahdat2021score}                      & Diffusion & Mixed Modeling              & \href{https://github.com/NVlabs/LSGM}{https://github.com/NVlabs/LSGM}                                                                                \\
2021 & Denoising diffusion GANs~\cite{xiao2021tackling} & Diffusion & Mixed Modeling              & \href{https://github.com/NVlabs/denoising-diffusion-gan}{https://github.com/NVlabs/denoising}                                          \\
2021 & DiffFlow~\cite{zhang2021diffusion}               & Diffusion & Mixed Modeling              & \href{https://github.com/qsh-zh/DiffFlow}{https://github.com/qsh-zh/DiffFlow}                                                                       

\\ \hline
    \end{tabular}}
    \caption{Major vision generative models.}
    \label{tab:unimodal_nlg}
\end{table}

\subsection{Vision Language}
\begin{table}[H]
    \centering
    \resizebox{0.9\textwidth}{!}{
    \begin{tabular}{lllll}
    \hline
    \textbf{Year} & \textbf{Method}                  & \textbf{Task}     & \textbf{Architecture}  & \textbf{Code}
     \\ \hline    
2019                              & VisualBERT~\cite{li2019visualbert}                       & VL Encoders       & Concatenated Encoders  & \href{https://github.com/uclanlp/visualbert}{https://github.com/uclanlp/visualbert}                   \\
2020                              & VL-BERT \cite{su2019vl}                         & VL Encoders       & Concatenated Encoders  & \href{https://github.com/jackroos/VL-BERT}{https://github.com/jackroos/VL-BERT}                       \\
2020                              & UNITER~\cite{zhou2020unified}                          & VL Encoders       & Concatenated Encoders  & \href{https://github.com/ChenRocks/UNITER}{https://github.com/ChenRocks/UNITER}                       \\
2021                              & ViLT~\cite{kim2021vilt}                             & VL Encoders       & Concatenated Encoders  & \href{https://github.com/dandelin/vilt}{https://github.com/dandelin/vilt}                             \\
2022                              & SimVLM~\cite{wang2021simvlm}                          & VL Encoders       & Concatenated Encoders  & -                                                                                                     \\
2019                              & LXMERT~\cite{tan2019lxmert}                          & VL Encoders       & Cross-aligned Encoders & \href{https://github.com/airsplay/lxmert}{https://github.com/airsplay/lxmert}                         \\
2019                              & 
X-LXMERT~\cite{cho2020x} & VL Encoders       & Cross-aligned Encoders & \href{https://github.com/allenai/x-lxmert}{https://github.com/allenai/x-lxmert}                       \\
2020                              & PixelBERT~\cite{huang2020pixel}                       & VL Encoders       & Concatenated Encoders  & \href{https://github.com/microsoft/xpretrain}{https://github.com/microsoft/xpretrain}                 \\
2019                              & ViLBERT~\cite{vilbert}                         & VL Encoders       & Cross-aligned Encoders & -                                                                                                     \\
2021                              & WenLan~\cite{huo2021wenlan}                          & VL Encoders       & Cross-aligned Encoders & \href{https://github.com/BAAI-WuDao/BriVl}{https://github.com/BAAI-WuDao/BriVl}                       \\
2021                              & CLIP~\cite{radford2019language}                           & VL Encoders       & Cross-aligned Encoders & \href{https://github.com/openai/CLIP}{https://github.com/openai/CLIP}                                 \\
2019                              & VLP~\cite{zhou2020unified}                              & To-text Decoders  & Encoder-Decoders       & \href{https://github.com/LuoweiZhou/VLP}{https://github.com/LuoweiZhou/VLP}                           \\
2021                              & ALBEF~\cite{li2021align}                            & To-text Decoders  & Encoder-Decoders       & \href{https://github.com/salesforce/ALBEF}{https://github.com/salesforce/ALBEF}                       \\
2022                              & BLIP~\cite{li2022blip}                             & To-text Decoders  & Encoder-Decoders       & \href{https://github.com/salesforce/lavis}{https://github.com/salesforce/lavis}                       \\
2023                              & BLIP-2~\cite{li2023blip}                           & To-text Decoders  & Frozen Decoders        & \href{https://github.com/salesforce/lavis}{https://github.com/salesforce/lavis}                       \\
2021                              & Frozen~\cite{tsimpoukelli2021multimodal}                           & To-text Decoders  & Frozen Decoders        & -                                                                                                     \\
2022                              & Flamingo~\cite{alayrac2022flamingo}                         & To-text Decoders  & Frozen Decoders        & \href{https://github.com/lucidrains/flamingo-pytorch}{https://github.com/lucidrains/flamingo-pytorch} \\
2023                              & Grounding~\cite{koh2023grounding}                        & To-text Decoders  & Frozen Decoders        & \href{https://github.com/kohjingyu/fromage}{https://github.com/kohjingyu/fromage}                     \\
2017                              & StackGAN~\cite{zhang2017stackgan}                         & To-image Decoders & GAN Decoders           & \href{https://github.com/hanzhanggit/StackGAN}{https://github.com/hanzhanggit/StackGAN}               \\
2018                              & AttnGAN~\cite{xu2018attngan}                          & To-image Decoders & GAN Decoders           & \href{https://github.com/taoxugit/AttnGAN}{https://github.com/taoxugit/AttnGAN}                       \\
2021                              & StyleCLIP~\cite{patashnik2021styleclip}                        & To-image Decoders & GAN Decoders           & \href{https://github.com/orpatashnik/StyleCLIP}{https://github.com/orpatashnik/StyleCLIP}             \\
2021                              & GLIDE~\cite{glide}                            & To-image Decoders & Diffusion Decoders     & \href{https://github.com/openai/glide-text2im}{https://github.com/openai/glide-text2im}               \\
2022                              & Stable-diffusion~\cite{rombach_high-resolution_2022}                 & To-image Decoders & Diffusion Decoders     & \href{https://github.com/compvis/stable-diffusion}{https://github.com/compvis/stable-diffusion}       \\
2022                              & Imagen~\cite{saharia2022photorealistic}                           & To-image Decoders & Diffusion Decoders     & \href{https://github.com/lucidrains/imagen-pytorch}{https://github.com/lucidrains/imagen-pytorch}     \\
2022                              & DALL-E-2~\cite{ramesh_hierarchical_2022}                         & To-image Decoders & Diffusion Decoders     & \href{https://github.com/lucidrains/DALLE2-pytorch}{https://github.com/lucidrains/DALLE2-pytorch}     \\
2021                              & DALL-E~\cite{ramesh2021zero}                           & To-image Decoders & VAE Decoders           & \href{https://github.com/openai/DALL-E}{https://github.com/openai/DALL-E}

\\ \hline
    \end{tabular}}
    \caption{Major vision language models.}
    \label{tab:multimodal_vl}
\end{table}

\subsection{Text Audio}
\begin{table}[H]
    \centering
    \resizebox{0.9\textwidth}{!}{
    \begin{tabular}{lllll}
    \hline
    \textbf{Year} & \textbf{Method}                    & \textbf{Task}         & \textbf{Code}             
     \\ \hline                                         2021 & AdaSpeech~\cite{chen2021adaspeech}       & Text-Audio Generation & \href{https://github.com/rishikksh20/AdaSpeech}{https://github.com/rishikksh20/AdaSpeech}                           \\
2021 & AdaSpeech2                               & Text-Audio Generation & \href{https://github.com/rishikksh20/AdaSpeech2}{https://github.com/rishikksh20/AdaSpeech2}                         \\
2020 & Lombard~\cite{paul2020enhancing}         & Text-Audio Generation & \href{https://github.com/dipjyoti92/TTS-Style-Transfer}{https://github.com/dipjyoti92/TTS-Style-Transfer}           \\
2019 & Zhang et al.\cite{zhang2019learning}     & Text-Audio Generation & \href{https://github.com/PaddlePaddle/PaddleSpeech}{https://github.com/PaddlePaddle/PaddleSpeech}                   \\
2019 & Yu et al.\cite{yu2019deep}               & Text-Music Generation & -                                                                                                                   \\
2018 & JTAV~\cite{liang2018jtav}                & Text-Music Generation & \href{https://github.com/mengshor/JTAV}{https://github.com/mengshor/JTAV}                                           \\
2021 & Ferraro et al.\cite{ferraro2021enriched} & Text-Music Generation & \href{https://github.com/andrebola/contrastive-mir-learning}{https://github.com/andrebola/contrastive-mir-learning} \\
2016 & Choi et al.\cite{choi2016towards}        & Text-Music Generation & -                                                                                                                   \\
2021 & MusCaps~\cite{manco2021muscaps}          & Text-Music Generation & \href{https://github.com/ilaria-manco/muscaps}{https://github.com/ilaria-manco/muscaps}                             \\
2022 & Manco et al.\cite{manco2022learning}     & Text-Music Generation & \href{https://github.com/ilaria-manco/mulap}{https://github.com/ilaria-manco/mulap}                                 \\
2022 & CLAP \cite{elizalde2022clap}             & Text-Music Generation & \href{https://github.com/YuanGongND/vocalsound}{https://github.com/YuanGongND/vocalsound}                           \\
2020 & Jukebox~\cite{dhariwal2020jukebox}       & Text-Music Generation & \href{https://github.com/openai/jukebox}{https://github.com/openai/jukebox}

\\ \hline
    \end{tabular}}
    \caption{Major text audio models.}
    \label{tab:multimodal_ta}
\end{table}

\subsection{Text Graph}
\begin{table}[H]
    \centering
    \resizebox{0.9\textwidth}{!}{
    \begin{tabular}{lllll}
    \hline
    \textbf{Year} & \textbf{Method}                    & \textbf{Task}         & \textbf{Code}             
     \\ \hline                
2016 & Li et al.~\cite{li2016commonsense}             & Text-to-KG Generation    & -                                                                                                                                                                 \\
2019 & KG-BERT~\cite{yao2019kg}                       & Text-to-KG Generation    & \href{https://github.com/yao8839836/kg-bert}{https://github.com/yao8839836/kg-bert}                                                                               \\
2020 & Malaviya et al.~\cite{malaviya2020commonsense} & Text-to-KG Generation    & \href{https://github.com/allenai/commonsense-kg-completion}{https://github.com/allenai/commonsense-kg-completion}                                                 \\
2019 & Petroni et al.~cite{petroni2019language}       & Text-to-KG Generation    & \href{https://github.com/facebookresearch/LAMA}{https://github.com/facebookresearch/LAMA}                                                                         \\
2020 & Shin et al.~cite{shin2020autoprompt}           & Text-to-KG Generation    & \href{https://github.com/ucinlp/autoprompt}{https://github.com/ucinlp/autoprompt}                                                                                 \\
2021 & Li et al.~cite{li2021prefix}                   & Text-to-KG Generation    & \href{https://github.com/XiangLi1999/PrefixTuning}{https://github.com/XiangLi1999/PrefixTuning}                                                                   \\
2022 & Lu et al.~\cite{lu2022unified}                 & Text-to-KG Generation    & \href{https://github.com/universal-ie/UIE}{https://github.com/universal-ie/UIE}                                                                                   \\
2022 & Grapher~\cite{melnyk2022knowledge}             & Text-to-KG Generation    & \href{https://github.com/ibm/grapher}{https://github.com/ibm/grapher}                                                                                             \\
2020 & CycleGT~\cite{2006.04702}                      & Text-KG Generation       & \href{https://github.com/QipengGuo/CycleGT}{https://github.com/QipengGuo/CycleGT}                                                                                 \\
2020 & DualTKB~\cite{2010.14660}                      & Text-KG Generation       & -                                                                                                                                                                 \\
2018 & GTR-LSTM~\cite{distiawan2018gtr}               & KG-to-Text Generation    & -                                                                                                                                                                 \\
2018 & Song et al.~\cite{1805.02473}                  & KG-to-Text Generation    & \href{https://github.com/freesunshine0316/neural-graph-to-seq-mp}{https://github.com/freesunshine0316/neural-graph-to-seq-mp}                                     \\
2020 & DUALENC~\cite{zhao-etal-2020-bridging}         & KG-to-Text Generation    & \href{https://github.com/zhaochaocs/DualEnc}{https://github.com/zhaochaocs/DualEnc}                                                                               \\
2019 & Koncel-Kedziorski et al.~\cite{koncel2019text} & KG-to-Text Generation    & \href{https://github.com/rikdz/GraphWriter}{https://github.com/rikdz/GraphWriter}                                                                                 \\
2020 & Ribeiro et al.~\cite{ribeiro2020modeling}      & KG-to-Text Generation    & \href{https://github.com/UKPLab/kg2text}{https://github.com/UKPLab/kg2text}                                                                                       \\
2020 & HetGT~\cite{yao2020heterogeneous}              & KG-to-Text Generation    & \href{https://github.com/QAQ-v/HetGT}{https://github.com/QAQ-v/HetGT}                                                                                             \\
2016 & Dong et al.~\cite{dong2016language}            & Semantic Parsing         & \href{https://github.com/donglixp/lang2logic}{https://github.com/donglixp/lang2logic}                                                                             \\
2016 & Jia et al.~\cite{jia2016data}                  & Semantic Parsing         & \href{https://worksheets.codalab.org/worksheets/0x50757a37779b485f89012e4ba03b6f4f}{https://worksheets.codalab.org/...} \\
2018 & Lyu et al.~\cite{lyu2018amr}                   & Semantic Parsing         & \href{https://github.com/ChunchuanLv/AMR_AS_GRAPH_PREDICTION}{https://github.com/...PREDICTION}                                             \\
2018 & Chen et al.~\cite{chen2018sequence}            & Semantic Parsing         & \href{https://github.com/dongpobeyond/Seq2Act}{https://github.com/dongpobeyond/Seq2Act}                                                                           \\
2019 & Zhang et al.~\cite{1905.08704}                 & Semantic Parsing         & \href{https://github.com/sheng-z/stog}{https://github.com/sheng-z/stog}                                                                                           \\
2019 & Fancellu et al.~\cite{fancellu2019semantic}    & Semantic Parsing         & -                                                                                                                                                                 \\
2021 & Text2Mol~\cite{inproceedings}                  & Text-Molecule Generation & \href{https://github.com/cnedwards/text2mol}{https://github.com/cnedwards/text2mol}                                                                               \\
2022 & MolT5~\cite{2204.11817}                        & Text-Molecule Generation & \href{https://github.com/blender-nlp/MolT5}{https://github.com/blender-nlp/MolT5}                                                                                 \\
2022 & MoMu~\cite{2209.05481}                         & Text-Molecule Generation & \href{https://github.com/bingsu12/momu}{https://github.com/bingsu12/momu}                                                                                        
     \\ \hline
    \end{tabular}}
    \caption{Major text graph models.}
    \label{tab:multimodal_ta}
\end{table}

\subsection{Text Code}
\begin{table}[H]
    \centering
    \resizebox{0.9\textwidth}{!}{
    \begin{tabular}{lllll}
    \hline
    \textbf{Year} & \textbf{Method}                    & \textbf{Task}         & \textbf{Code}             
     \\ \hline        
     2020 & CodeBERT~\cite{feng2020codebert} & Text-Code Generation & \href{https://github.com/microsoft/CodeBERT}{https://github.com/microsoft/CodeBERT}                                                                 \\
2020 & CodeBERT~\cite{feng2020codebert}   & Text-Code Generation & \href{https://github.com/microsoft/CodeBERT}{https://github.com/microsoft/CodeBERT}                                                                 \\
2020 & CuBERT~\cite{kanade2020learning}   & Text-Code Generation & \href{https://github.com/google-research/google-research/tree/master/cubert}{https://github.com/google-research/google-research/tree/master/cubert} \\
2021 & CodeT5~\cite{wang2021codet5}       & Text-Code Generation & \href{https://github.com/salesforce/codet5}{https://github.com/salesforce/codet5}                                                                   \\
2021 & PLBART~\cite{ahmad2021unified}     & Text-Code Generation & \href{https://github.com/wasiahmad/PLBART}{https://github.com/wasiahmad/PLBART}                                                                     \\
2017 & Yin et al.~\cite{yin2017syntactic} & Text-Code Generation & \href{https://github.com/pcyin/NL2code}{https://github.com/pcyin/NL2code}                                                                           \\
2018 & Dai et al.~\cite{dai2018syntax}    & Text-Code Generation & \href{https://github.com/Hanjun-Dai/sdvae}{https://github.com/Hanjun-Dai/sdvae}                                                                     \\
2022 & CODEGEN~\cite{2203.13474}          & Text-Code Generation & \href{https://github.com/salesforce/CodeGen}{https://github.com/salesforce/CodeGen}                                                                 \\
2022 & TDUIF~\cite{2208.05950}            & Text-Code Generation & -                                                                                                                                                  

     \\ \hline
    \end{tabular}}
    \caption{Major text code models.}
    \label{tab:multimodal_ta}
\end{table}

\end{document}